%% file: main.tex
\documentclass[journal]{IEEEtran}

\usepackage{amsmath,epsfig}
\usepackage{amsfonts}
\usepackage{times}
\usepackage{balance}
\usepackage{amssymb}
\usepackage{graphicx}
\usepackage{euscript}
\usepackage{algorithm, algorithmic}
\usepackage{color}
\usepackage{breqn}
\usepackage{subfigure}
\usepackage{multirow,hhline}
\usepackage{tabularx,colortbl}
\usepackage{placeins}
\usepackage{balance}

\newcommand{\ru}    {\rule{0mm}{4mm}}

\begin{document}

\title{Media Forensics and DeepFakes: \\an overview}

\author{Luisa Verdoliva% <-this % stops a space

\IEEEcompsocitemizethanks{
\IEEEcompsocthanksitem L.Verdoliva is with the Department of Industrial Engineering, Universit\`{a} Federico II di Napoli, Naples, Italy. E-mail: verdoliv@unina.it.}}

\maketitle

\begin{abstract}
	With the rapid progress of recent years,
	techniques that generate and manipulate multimedia content can now guarantee a very advanced level of realism.
	The boundary between real and synthetic media has become very thin.
	On the one hand, this opens the door to a series of exciting applications in different fields such as creative arts, advertising, film production, video games.
	On the other hand, it poses enormous security threats.
	Software packages freely available on the web allow any individual, without special skills, to create very realistic fake images and videos.
	So-called deepfakes can be used to manipulate public opinion during elections, commit fraud, discredit or blackmail people.
	Potential abuses are limited only by human imagination.
	Therefore, there is an urgent need for automated tools capable of detecting false multimedia content and avoiding the spread of dangerous false information.
	This review paper aims to present an analysis of the methods for visual media integrity verification, that is, the detection of manipulated images and videos.
	Special emphasis will be placed on the emerging phenomenon of deepfakes and, from the point of view of the forensic analyst, on modern data-driven forensic methods.
	The analysis will help to highlight the limits of current forensic tools, the most relevant issues, the upcoming challenges,
	and suggest future directions for research.
\end{abstract}


\input{introduction}
\input{fake_content}
\input{data_integrity}
\input{deep_learning}
\input{deepfakes}
\input{considerations}
\input{datasets}
\input{counterforensics}
\input{fusion}
\input{future_work}

\section{Conclusion}

Fifteen years ago multimedia forensics was a niche field
of practical interest only for a restricted set of players involved in law enforcement, intelligence, private investigations.
Both attacks and defences had an artisan flavor, and required painstaking work and dedication.

Artificial intelligence has largely changed these rules.
High-quality fakes now seem to come out from an assembly line calling for an extraordinary effort on part of both scientists and policymakers.
In fact, today's multimedia forensics is in full development,
major agencies are funding large research initiatives, and scientists form many different field are contributing actively, with fast advances in ideas and tools.

It is difficult to forecast whether such efforts will be able to ensure information integrity in the future, or some forms of active protection will become necessary.
This is an arms race, and one part is no smarter than the other.
For the present time,
a large arsenal of tools is being developed,
and knowing them, the principles on which they rely, and their scope of application is a prerequisite to protect institutions and ordinary people.

\section{Acknowledgement}

We gratefully acknowledge the support of this research by a Google Faculty Award. 
In addition, this material is based on research sponsored by the Air Force Research
Laboratory and the Defense Advanced Research Projects Agency under agreement number FA8750-16-2-0204. 
The U.S. Government is authorized to reproduce and distribute reprints for Governmental purposes 
notwithstanding any copyright notation thereon. 
The views and conclusions contained herein are those of the authors and should not be interpreted
as necessarily representing the official policies or endorsements, 
either expressed or implied, of the Air Force Research Laboratory and the Defense Advanced Research Projects Agency or the U.S. Government.

\balance

\ifCLASSOPTIONcaptionsoff
  \newpage
\fi

\bibliographystyle{IEEEtran}
\bibliography{refs}

\end{document}

%% file: introduction.tex
\section{Introduction}
\label{sec:introduction}

Fake multimedia has become a central problem in the last few years,
especially after the advent of the so called {\em Deepfakes},
i.e., images and videos manipulated using advanced deep learning tools,
like autoencoders (AE) or generative adversarial networks (GAN).
With this technology, creating realistic manipulated media assets may be very easy,
provided one can access large amounts of data.
Applications include movie productions, photography, video-games and virtual reality.
The very same technology, however, can also be used for malicious purposes,
like creating fake porn videos to blackmail people, or building fake-news campaigns to manipulate the public opinion.
In the long run, it may also reduce trust in journalism, including serious and reliable sources.
Figure 1 shows some popular deepfakes circulating on the internet.
These fakes are easy to spot since they were generated for fun and involve well-known actors and politicians in unlikely situations.
In addition, on the web it is usually possible to retrieve both the original and the manipulated version,
removing any doubt about authenticity.
However, verifying digital integrity becomes much more difficult
if the video portrays a less known person and only the manipulated version is publicly available.
This scenario takes place, for example, if the attacker films a new video on his own,
with a collaborative actor whose face is eventually replaced by the target face.
Governmental bodies, enforcement agencies, the news industry, and also the man in the street are becoming acutely aware of the potential menace carried by such a technology.
The scientific community is asked to develop reliable tools for automatically detecting fake multimedia.

Actually, this is not a new problem.
Image manipulation has been carried out since photography was born\footnote{https://www.dailymail.co.uk/news/article-2107109/Iconic-Abraham-Lincoln-portrait-revealed-TWO-pictures-stitched-together.html},
and powerful image/video editing tools, such as Photoshop\textregistered, After Effects Pro\textregistered, or the open source software GIMP, have been around for a long time.
Using such conventional signal processing methods,
images can be easily modified, obtaining realistic results that can fool even a careful observer.
Figure 2 shows some examples of skillfully manipulated images
that have been disseminated on the Internet in recent years to spread false news, both on images\footnote{https://www.cnn.com/2018/03/26/us/emma-gonzalez-photo-doctored-trnd/index.html}
and videos\footnote{https://www.theguardian.com/world/2015/mar/19/i-faked-the-yanis-varoufakis-middle-finger-video-says-german-tv-presenter}.
In fact, research in multimedia forensics has been going on for at least 15 years \cite{Farid2009a, Farid2016},
and is receiving ever growing attention, not only from the academy, but also from major information technology (IT) companies and funding agencies.
In 2016, the Defense Advanced Research Projects Agency (DARPA) of the U.S. Department of Defense
launched the large-scale Media Forensic initiative (MediFor) to foster research on media integrity,
with important outcomes in terms of methods and reference datasets.

\begin{figure}
	\centering
	\includegraphics[width=0.95\linewidth]{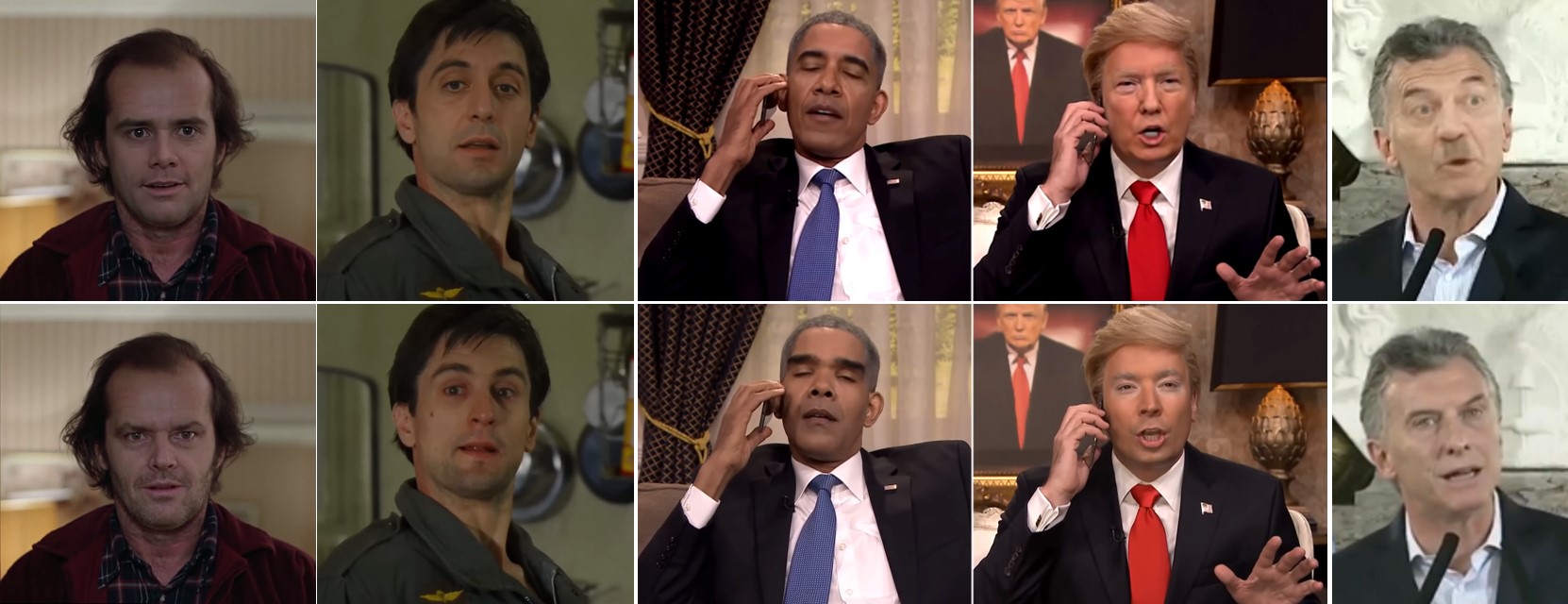}
	\caption{Examples of deepfake manipulations from YouTube. Top: manipulated videos; bottom: original videos. It is worth noting that in the real videos Obama and Trump are impersonated by comic actors.}
	\label{fig:deepfakes}
\end{figure}

\begin{figure}[t!]
	\centering
	\begin{tabular}{cc}
		\includegraphics[width=1.50\linewidth, trim=40 150 0 60, clip]{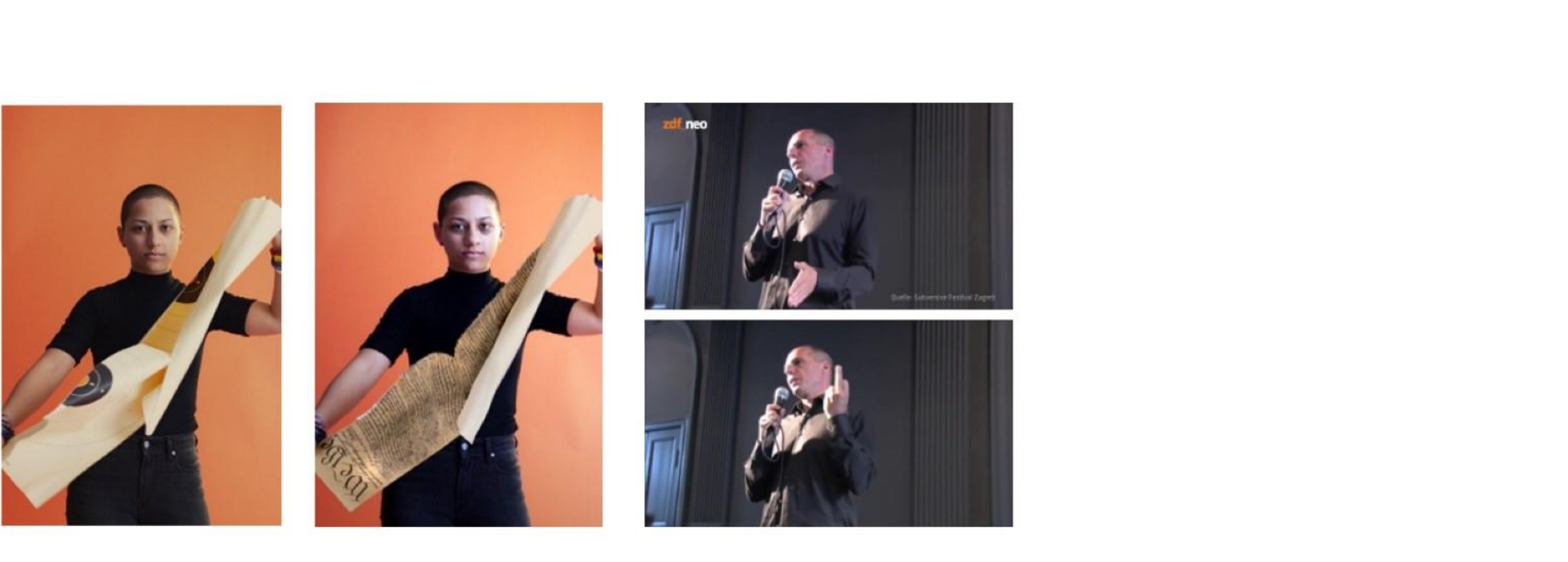}
	\end{tabular}
	\caption{Examples of fake multimedia where two different versions of an image/video can be retrieved from the web. In one image Emma Gonzalez (left) an american activist is tearing up the American Constitution, while in a video Yanis Varoufakis (right) a greek politician is giving the middle-finger gesture to Germany.}
	\label{fig:real_cases}
\end{figure}

Following the MediFor taxonomy,
digital media verification should look for physical integrity, digital integrity, and semantic integrity.
In the literature, several methods have been proposed,
which expose physical inconsistencies, concerning for example shadows or illumination or perspective \cite{Johnson2007, Kee2013, Carvalho2013}.
Modern sophisticated manipulations, however, are more and more effective in avoiding such pitfalls and 
methods which test digital integrity are by far more widespread and represent the current state of the art.
Indeed, each image or video is characterized by a number of features,
which depend on the different phases of its digital history:
from the very same acquisition process, to the internal camera processing (e.g. demosaicing, compression),
to all external processing and editing operations \cite{Piva2012}.
Digital manipulations tend to modify such features,
leaving a trail of clues which, although invisible to the eye, can be exploited by pixel-level analysis tools.
Instead, semantic integrity is violated when the media asset under analysis conveys information
which is not coherent with the context or with evidence coming from correlated sources.
For example, when objects are copy-pasted from images available on the web,
several near-identical copies can be detected \cite{Wu2017, Liu2019}, suggesting a possible manipulation.
Moreover, by identifying the connections among the various versions of the same asset,
it is possible to build its manipulation history (image and video phylogeny) \cite{Dias2011, Moreira2018}.

Despite the continuous research efforts and the numerous forensic tools developed in the past,
the advent of deep learning, 
is changing the rules of the game and asking multimedia forensics for new and timely solutions.
This phenomenon is also causing a strong acceleration in multimedia forensics research, 
which often relies itself on deep learning.
There have been several reviews on this topic \cite{Rocha2011, Milani2012, Piva2012, Stamm2013, Korus2017}, 
however these last years have witnessed the advent of new methods.
Hence, beyond reviewing the conventional media forensics approaches,
a special attention will be devoted to deep learning-based approaches and to the strategies designed to fight deepfakes.
The analysis will be restricted to passive methods and visual data-based solutions.
That is, it will be assumed that no active strategy is in place to ensure integrity,
and that a skilled attacker
modified metadata to make them useless,
otherwise they would provide precious information towards authenticity verification both for images and videos \cite{Kee2011, Iuliani2018, Guera2019}.
On the other hand, it is worth noting that metadata are routinely canceled when media assets are uploaded on a social network.

\begin{figure}[t!]
	\centering
	\begin{tabular}{cc}
		\includegraphics[width=2.0\linewidth, trim=0 230 70 45, clip]{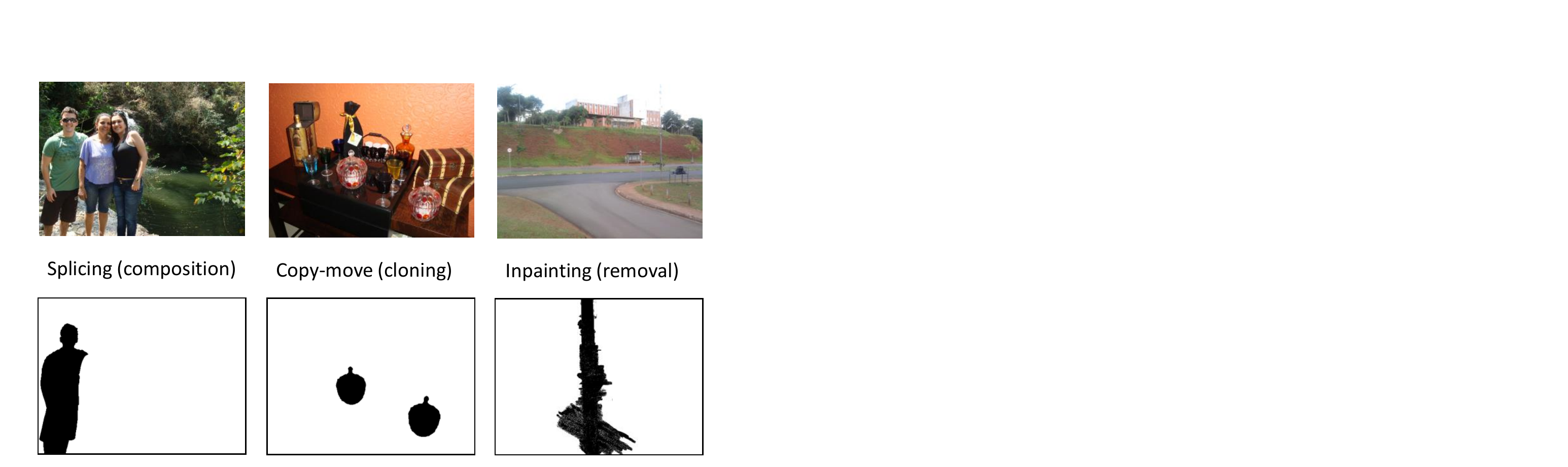}
	\end{tabular}
	\caption{Examples of image manipulations carried out using conventional media editing tools.
    Images come from the dataset of the 1st IEEE Image Forensics Challenge organized in 2013.
    From left to right:
    splicing (alien material has been inserted in the image),
    copy-move (an object has been cloned),
    inpainting (an object has been hidden by background patches).}
	\label{fig:trad_methods}
\end{figure}

The review starts with a brief analysis of the most effective manipulation methods proposed in recent years (Section II). Then, integrity verification methods are described,
beginning with conventional approaches (Section III),
then moving to deep learning-based approaches (Section IV),
to conclude with specific deepfake detection methods (Section V).
In Section  VI, a discussion of the state of multimedia forensics and its perspectives after the advent of deep learning is carried out.
A list of the datasets most widespread in the field is presented in Section VII.
Then, the further major themes of counterforensics (Section VIII) and fusion (Section IX) are considered.
Finally, future research directions are outlined (Section X) and conclusions are drawn (Section XI).

%% file: fake_content.tex
\begin{figure*}[t!]
	\centering
	\begin{tabular}{cc}
       \includegraphics[width=1.05\linewidth, trim=20 10 0 10, clip]{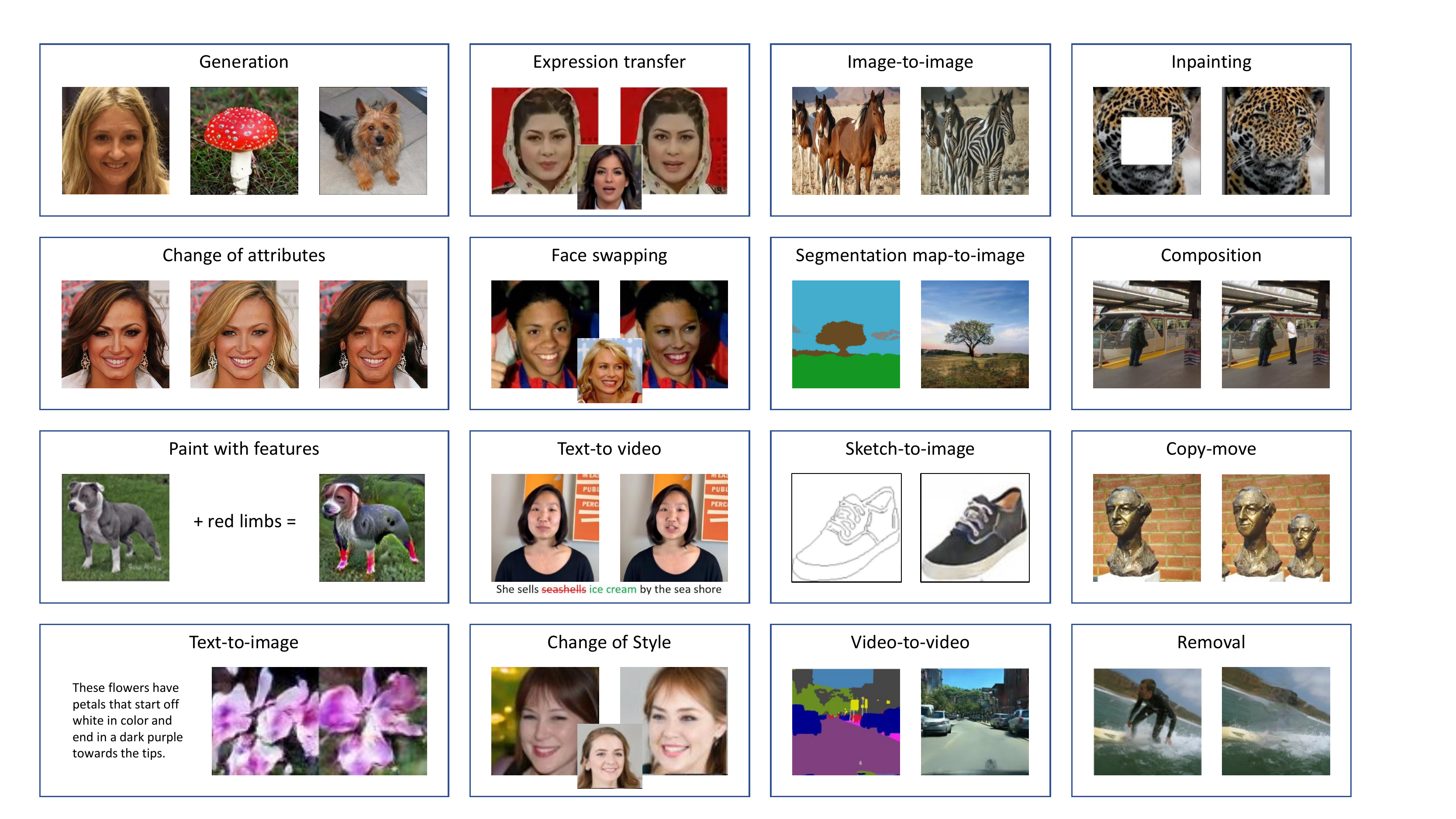}
	\end{tabular}
	\caption{Examples of image and video manipulations carried out using deep learning methods. 	
    Besides conventional manipulations, like composition \cite{Tan2018}, copy-move \cite{Thies2019}, object removal \cite{Shetty2018}, inpainting \cite{Yang2017}, a large number of new tasks can be performed.
    These include content generation \cite{Karras2019, Brock2019}, image/video synthesis from semantic labels \cite{Park2019, Wang2019synthesis} or sketches \cite{Isola2017} or text \cite{Reed2016}, changes of style and attributes \cite{Karras2019, Wu2019RelGAN, Engstrom2019}, domain translation \cite{Zhu2017}, up to expression transfer \cite{Wang2019synthesis}, face swapping typical of deepfakes \cite{Nirkin2018} and talking-head video editing \cite{Fried2019}.
    It is worth underlining that deep learning methods require no manual media editing on the part of the user, except for possible post-processing.
    }
	\label{fig:list_of_fakes}
\end{figure*}

\section{Fake content generation}

There are many ways to manipulate visual content, and new methods are proposed by the day.
This Section will briefly review some of the most widespread and promising of them.
Very common operations are adding, replicating or removing objects, as in the examples of Figure 3.
A new object can be inserted by copying it from a different image (splicing), or from the same image (copy-move).
Instead, an existing object can be deleted by extending the background to cover it (inpainting) like in the popular exemplar-based inpainting \cite{Barnes2009}.
All these tasks are easily accomplished with widespread image editing packages.
Then, some suitable post-processing, like resizing, rotation or color adjustment,
may be required to better fit the object to the scene, both to improve the visual appearance and to guarantee coherent perspective and scale.
In recent years, however, the same results are achieved, with better semantic consistency,
through advanced computer graphics (CG) approaches and deep learning  (see Figure \ref{fig:list_of_fakes}, last column).
Manipulations that do not require sophisticated artificial intelligence (AI) tools
are sometimes referred to as ``cheap fakes''.
Nonetheless, their impact in distorting reality can be very high.
For example, by removing, inserting or cloning entire groups of frames one can completely change the meaning of a video.
A simple frame-rate reduction was recently used to let Nancy Pelosi, Speaker of the U.S. House of Representatives,
appear as drunk or confused\footnote{https://www.washingtonpost.com/technology/\-2019/05/23/faked-pelosi-videos-slowed-make-her-appear-drunk-spread-across-social-media/}.

Besides these ``traditional'' manipulations, concerning specific areas of the image or video,
deep learning and computer graphics are now offering a large number of new ones.
First of all, a media asset can be synthesized completely from scratch.
To this end, autoencoders and generative adversarial networks allowed to develop successful solutions \cite{Huang2018} especially for face synthesis,  where a high level of photo-realism has been achieved \cite{Karras2018, Karras2019}.
It is also possible to generate a completely synthetic image or video using a segmentation map as input \cite{Wang2018}.
Image synthesis is also achievable using only a sketch \cite{Zakharov2019, Park2019} or a text description \cite{Reed2016}.
Likewise, the face of a person can be animated based on an audio input sequence \cite{Suwajanakorn2017, Chung2017}.
More often, the manipulation modifies existing images or videos.
A well-known example is style transfer \cite{Isola2017, Zhu2017}, which allows to change the style of a painting, switch oranges to apples, or reproduce an image in a different season.
Major efforts have been devoted to manipulating faces, for their high semantic value, and for the many possible applications.
Methods have been proposed to change the expression of a face \cite{Choi2018, Qian2019},
to transfer the expression from a source to a target actor \cite{Thies2016, Kim2018},
or to swap faces \cite{Nirkin2018}.
Recently, it has been shown that effective face manipulation is feasible even without a huge amount of training photos of the targeted person \cite{Nirkin2019}.
It is even possible to animate the face of a still portrait and express various types of emotions \cite{Elor2017}.
Beyond faces, some recent work addressed motion transfer:
the target person dances following the movements transferred from a source dancer \cite{Chan2019}.
In Figure 4 some examples of such manipulations are presented. One can easily observe how realistic they appear
and the variety of possible automatic editing tools available nowadays.

%% file: data_integrity.tex
\begin{figure*}[t!]
	\centering
	\begin{tabular}{cc}
		\includegraphics[width=1.00\linewidth, trim=20 190 0 80, clip]{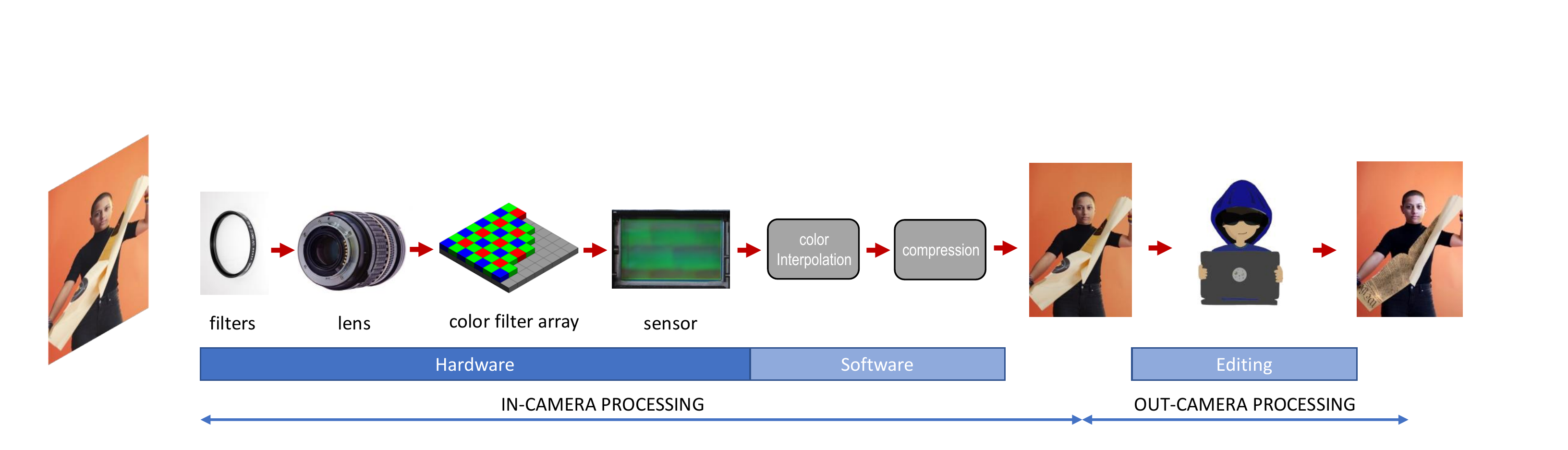}
	\end{tabular}
	\caption{An image is captured using an acquisition system whose basic components are represented in this figure.
		After undesired light components are reduced using optical filters, the lenses focus the light on the sensor.
        In order to extract the red-green-blue (RGB) components, a color filter array (CFA) is present.
		Each individual sensor element records light only in a certain range of wavelengths.
		Therefore, the missing color information at a pixel must be recovered from surrounding pixels,
        through a process known as color filter array interpolation or demosaicing.
        Then, a sequence of internal processing steps follow, including color correction, enhancement and, finally, compression.
		The implementation and parametrization of all these components differ based on the camera model
		and provide important clues that are exploited in the forensic multimedia analysis.
        Alterations carried out by the malicious user can also introduce artifacts that allow forensic analyses and detection.}
	\label{fig:acquisition}
\end{figure*}

\section{Conventional detection methods}

This Section reviews the major lines of research in multimedia forensics before the emergence of deep learning and deepfakes.
The most popular approaches look for artifacts
related to the in-camera processing chain (camera-based clues) or the out-camera processing history (editing-based clues) \cite{Swaminathan2008}.
A defining property of the approaches proposed so far is the prior knowledge they rely upon,
which impacts on their suitability for real-world applications.
Following this perspective,
first, blind methods will be described, where no prior knowledge is required.
Then, the focus will shift on one-class methods, which need information only on pristine data,
through a collection of images/videos taken from the camera of interest or, more in general, a large set of untampered data.
Eventually, supervised methods will be considered,
which rely on a suitable training set comprising both pristine and manipulated data.

\subsection{Blind methods}

\begin{figure*}[t!]
	\centering
	\begin{tabular}{cc}
		\includegraphics[width=1.05\linewidth, trim=20 320 0 10, clip]{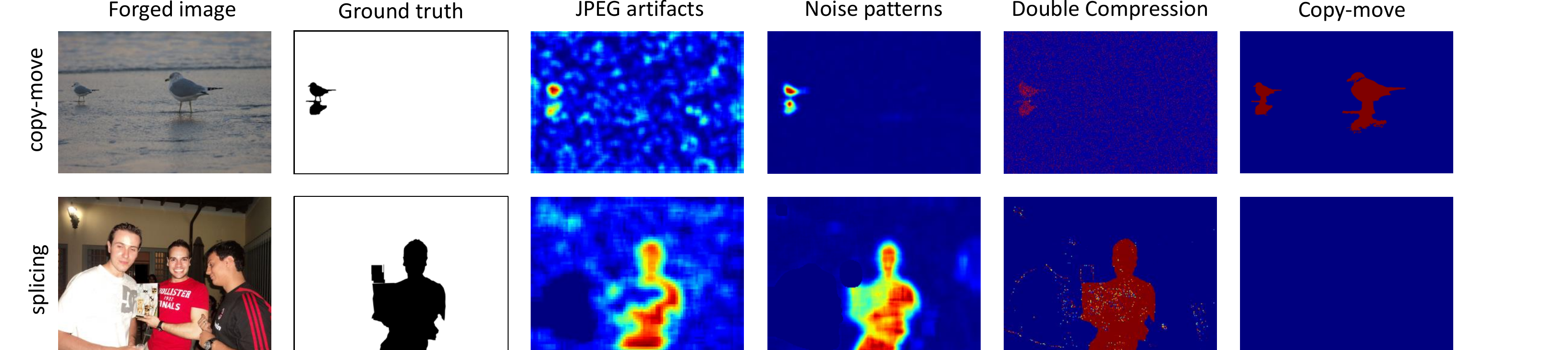}
	\end{tabular}
	\caption{Localization results of some blind methods for images with copy-move (top) and splicing (bottom).
    From left to right, manipulated image, ground truth, and localization heatmaps obtained with methods based on
    JPEG artifacts, noise patterns, double quantization artifacts, copy-move search. Of course, copy-move methods are not effective for splicing manipulations.}
	\label{fig:blind_uncompressed}
\end{figure*}

Blind approaches do not use any external data for training or for other forms of pre-processing:
they rely exclusively on the media asset under analysis, and try to reveal anomalies which may suggest the presence of an manipulation.
In particular, they look for a number of specific artifacts originated by in-camera or out-camera processing (Figure \ref{fig:acquisition}).
In fact, the image formation process inside a camera requires a number of operations, both hardware and software,
which are specific of each individual camera and leave distinctive traces on the acquired image.
For example, the demosaicing algorithm is typically different for different camera models.
Therefore, when a manipulation involves the composition of parts of images acquired from different models, demosaicing-related spatial anomalies arise.
Likewise, the out-camera editing process may introduce its own peculiar traces, as well as disrupt fingerprint-like camera-specific patterns,
phenomena which both allow reliable detection of the attack.
Of course, most of these traces are very subtle and cannot be perceived at a visual inspection.
However, once properly emphasized, they represent a precious source of information to establish digital integrity (Figure \ref{fig:blind_uncompressed}).

\subsubsection{Lens distortion}
each camera is equipped with a complex optical system which cannot perfectly focus light at all different wavelengths.
These imperfections can be used for forensic purposes.
In \cite{Johnson2006} a method is proposed which exploits the lateral chromatic aberrations,
off-axis displacements of the light components at different wavelengths that results in a misalignment between the color channels, while the method proposed in \cite{Yerushalmy2011}
relies on the aberrations generated by the interaction between lens and sensor.
Improved versions of the method based on lateral chromatic aberrations, 
with a more efficient estimation of local displacements,
are proposed in \cite{Gloe2010} and more recently in \cite{Mayer2018a}.
Finally, in \cite{Fu2012} it is exploited the radial distortion that characterizes the wide-angle lens typically used for indoor/outdoor video surveillance.

\subsubsection{CFA artifacts}
most digital cameras use a color filter array (CFA), with a periodic pattern,
so that each individual sensor element records light only in a certain range of wavelengths (i.e. red, green, blue).
The missing color information is then interpolated from surrounding pixels, an operation known as demosaicing.
This process introduces a subtle periodic correlation pattern in all acquired images.
Whenever a manipulation occurs, this periodic pattern is perturbed.
In addition, since CFA configuration and interpolation algorithms are specific of each camera model \cite{Bayram2005, Cao2009},
when a region is spliced in a photo taken by another camera model, its periodic pattern will appear anomalous.
One of the first methods to exploit these artifacts was proposed by Popescu and Farid \cite{Popescu2005} back in 2005,
based on a simple linear model to capture periodic correlations.
Of course, periodic signals produce strong peaks in the Fourier domain.
This can be used to distinguish natural images from computer generated images \cite{Popescu2005, Gallagher2008},
especially after high-pass filtering the image so as to extract more effective features \cite{Gallagher2008, Dirik2009}.
The problem can be also recast in a Bayesian framework, as proposed in \cite{Ferrara2012},
obtaining a probability map in output which allows for fine-grained localization of image tampering.
In \cite{Ho2010} the analysis is extended to take into account also pixel correlations across color channels.

\subsubsection{Noise level and noise pattern}
a more general approach is to highlight noise artifacts introduced by the whole acquisition process, irrespective of their specific origin.
The analysis of {\em local} noise level may help reveal splicings, as shown in \cite{Popescu2004},
because different cameras are characterized by different intrinsic noise.
Local noise analysis has been proposed using statistical tools in the image domain \cite{Popescu2004, Lyu2014} or in the wavelet domain \cite{Mahdian2009}.
This approach is also at the basis of the so-called Error Level Analysis (ELA), widely used by practitioners for its simplicity.
However, noise intensity alone is not very informative, and may easily provide wrong indications.
Therefore, in \cite{Cozzolino2015} the high-pass noise residual of the image is used to extract rich features which better characterize local neighborhoods.
The expectation-maximization algorithm is then used for clustering these features and reveal possible anomalies.
In all above methods, the noise residual is only used to detect possible anomalies.

Departing from this ``agnostic'' approach,
the method proposed in \cite{Swaminathan2008} uses the noise residual to estimate the imaging model and define an intrinsic camera fingerprint.
Inconsistencies with respect to the estimated model are then used to discover possible manipulations.
A similar idea is extended to videos in \cite{Hsu2008, Mullan2017, Ding2018} where the noise residuals of consecutive frames are analyzed
and suitable features are extracted to discover traces of both intra-frame and inter-frame manipulations.
Instead, in \cite{Kobayashi2010}, the camera dependent photon shot noise is considered as an alternative fingerprint for static scenes.

\subsubsection{Compression artifacts}
exploiting compression artifacts, has long been a workhorse in image forensics.
The many methods proposed in the literature, mostly for JPEG compressed images, can be classified based on the clues they rely upon.
A first popular approach is to exploit the so-called lock artifact grid (BAG).
Because of the block-wise JPEG processing, discontinuities appear along the block boundaries of compressed images,
giving rise to a distinctive and easily detected grid-like pattern \cite{Fan2003}.
In the presence of splicing or copy-move manipulations, the BAGs of inserted object and host image typically mismatch, enabling detection.
Several BAG-based methods have been proposed in the literature \cite{Luo2007, Li2009, Lin2009},
some of which even in recent years \cite{Iakovidou2018}.

Another major approach relies on double compression traces.
In fact, when a JPEG-compressed image undergoes a local manipulation and is compressed again,
double compression artifacts appear all over the image except in the forged region \cite{Lukas2003}.
These artifacts change depending on whether the two compressions are spatially aligned or not,
but suitable detection \cite{Chen2011} and localization \cite{Barni2010, Bianchi2012} methods have been proposed for both cases.
Another method relies on the so-called JPEG ghosts \cite{Farid2009b},
arising in the manipulated area when two JPEG compressions use the same quality factor (QF).
To highlight ghosts, the target image is compressed at all QFs and analyzed.
Other methods \cite{Fu2007, Pasquini2017} look for anomalies in the statistical distribution of original DCT samples,
assumed to comply with the Benford law.

A further approach is to exploit the model-specific implementations of the JPEG standard,
including customized quantization tables and post-processing steps \cite{Farid2006, Kee2011}.
In \cite{Agarwal2017} model-specific JPEG features have been defined, the JPEG dimples,
which depend on how coefficients are converted from real to integer: by the ceil, floor, or rounding operator.
Also, chroma subsampling presents specific clues due to integer rounding \cite{Lorch2019}.

Exploiting compression artifacts for detecting video manipulation is also possible,
but is much more difficult because of the complexity of the video coding algorithm.
Traces of MPEG double compression were first highlighted in the seminal paper by Wang and Farid for detecting frames removal \cite{Wang2006}. In fact, the de-syncronization caused by removing a group of frames introduces spikes in the Fourier Transform of the motion vectors.
A successive work \cite{Padin2012} tried to improve the double compression estimation especially in the more challenging scenario when the strength of the second compression increases and 
proposed a distinctive footprint, based on the variation of the macroblock prediction types in the reencoded P-frames. This same artifact is exploited in \cite{Gironi2014} where its estimation is improved to detect traces of inter-frame tampering, and more recently in \cite{Padin2019} to deal with video
sequences that contain bi-directional frames.

\subsubsection{Editing artifacts}
the manipulation process often generates a trail of precious traces, besides artifacts related to re-compression.
Indeed, when a new object is inserted in an image, it typically needs several post-processing steps to fit well the new context.
These include geometric transformations, like rotation and scaling, contrast adjustment,
but also blurring, to smooth the object-background boundaries.
Therefore, many papers focus on detecting these basic operations as a proxy for possible forgeries.
Some methods \cite{Popescu2004, Kirchner2008} try to detect traces of resampling, always necessary in the presence of rotation or resizing by exploiting periodic artifacts.
Other approaches focus on anomalies on the boundaries of objects when a composition is performed \cite{Dong2006}, or by blur inconsistencies \cite{Bahrami2015}.

\begin{figure}[t!]
	\centering
	\begin{tabular}{cc}
		\includegraphics[width=1.15\linewidth, trim=80 30 0 90, clip]{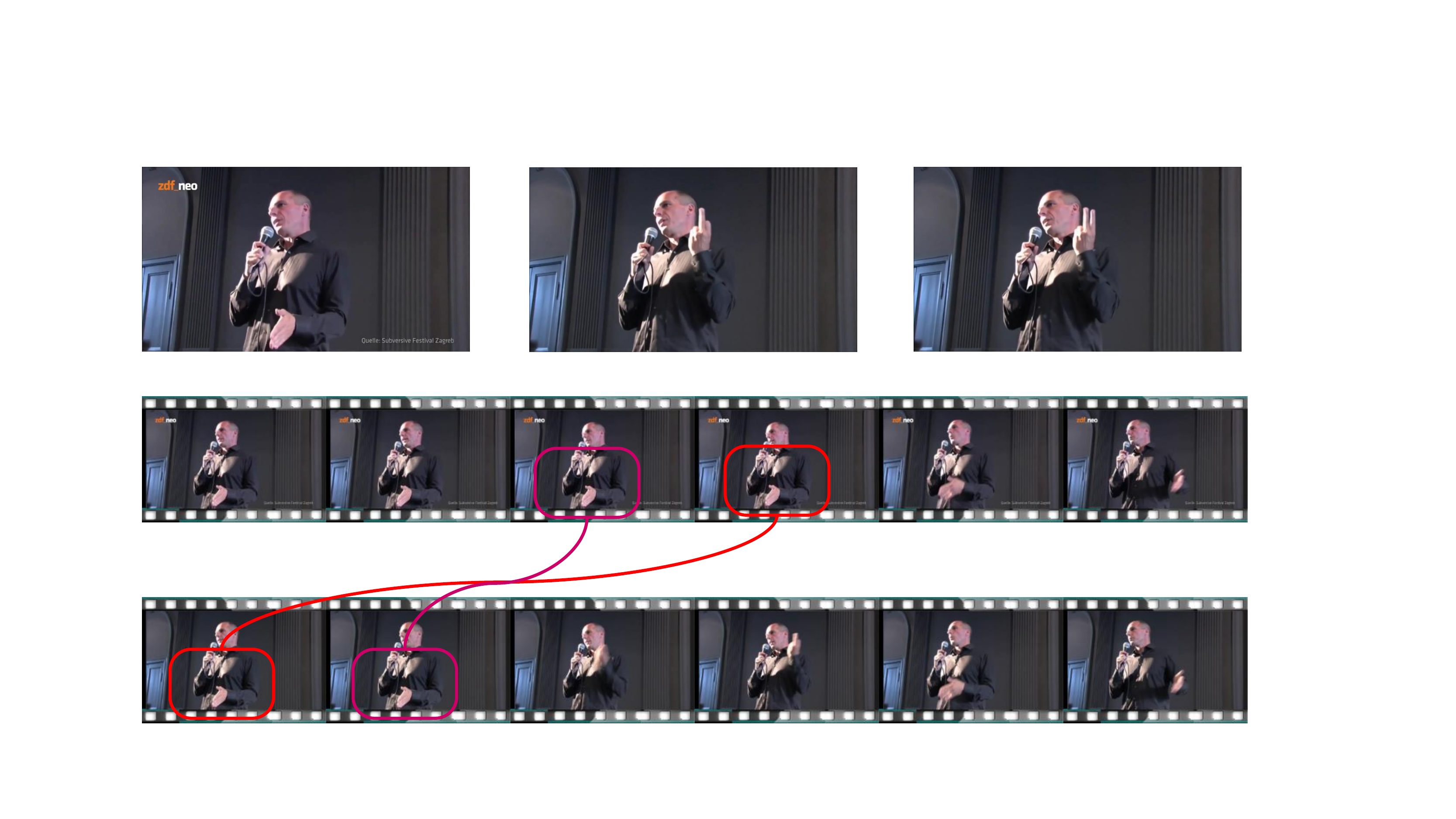}
	\end{tabular}
	\caption{The Varoufakis video copy-move.
    Three versions of the same video appeared on the web, with radically different content, as per the sample frames shown on the top row.
    With reference to the video on the left where the arm is down, a copy-move detector 
    %proposed in \cite{DAmiano2019} 
    revealed that a sequence from one video (second row) was
    temporally flipped and copy-moved onto another one (bottom row).
    For the other two videos the detector was not able to tell apart the pristine from the forged one, since the discriminative region was too small.}
	\label{fig:Varoufakis}
\end{figure}

A very common manipulation consists in copy-moving image regions to duplicate or hide objects.
Of course, the presence of identical regions is a strong hint of forgery,
but clones are often modified to disguise traces, and near-identical natural objects also exist, which complicate the forensic analysis.
Studies on copy-move detection date back to 2003, with the seminal work of Fridrich \cite{Fridrich2003}.
Since then, a large literature has grown on this topic.
Effective and efficient solutions are now available which allow for copy-move detection
even in the presence of rotation, resizing, and other geometric distortions \cite{Christlein2012}.
Methods based on keypoints \cite{Amerini2011, Silva2015} are very efficient,
while dense-field methods \cite{Ryu2013, Cozzolino2015efficient} are more accurate and deal also with occlusive attacks.
In \cite{Cozzolino2014b} dense-field methods have been shown to be effective also to detect inpainting.
Extensions to video have been also proposed both for detection and localization \cite{Bestagini2013, DAmiano2019}, the main issue being complexity.
In the example of Figure \ref{fig:Varoufakis}, a method for 3D copy-move localization 
exposed a copy-move with flipping in one of the videos.
As for inter-frame forgeries, local modifications can be detected based on the consistency of the velocity field \cite{Wu2014video}.

\subsection{One-class sensor-based and model-based methods}

The camera sensor can provide a wealth of precious clues.
In fact, due to manufacturing imperfections, the sensor elements present small deviations from their expected behavior.
Such deviations form a noise-like pattern, stable in time, called photo-response non-uniformity (PRNU) noise.
All images acquired by a given camera bear traces of its PRNU pattern,
which can be therefore regarded as a sort of camera fingerprint.
If a region of the image is tampered with, the corresponding PRNU pattern is removed, which allows one to detect the manipulation.

\begin{figure}[t!]
	\centering
	\begin{tabular}{cc}
		\includegraphics[width=1.05\linewidth, trim=20 10 0 30, clip]{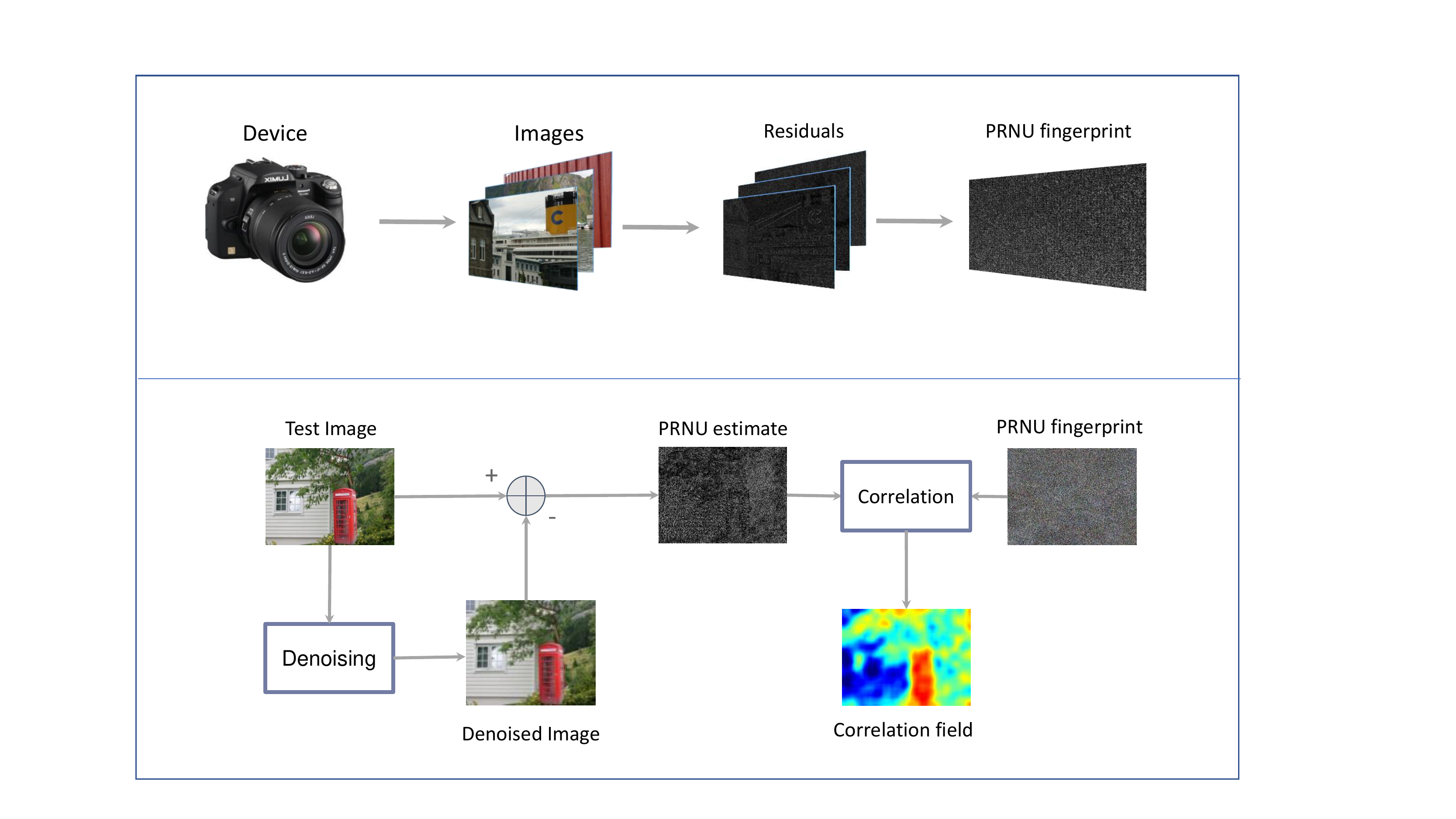}
	\end{tabular}
	\caption{PRNU-based forgery localization.
    Top: the device PRNU pattern is estimated by averaging a large number of noise residuals.
    Bottom: the image PRNU pattern is estimated through denoising, and compared with the reference pattern:
    the low correlation in the telephone booth region suggests a possible manipulation.}
	\label{fig:PRNU}
\end{figure}

PRNU-based forgery detection was first proposed in \cite{Lukas2006} based on two steps:
{\it  i)} the camera PRNU pattern is estimated off-line from a large number of images taken from the camera, and
{\it ii)} the target image PRNU is estimated at test time, by means of a denoising filter, and compared with the reference (see Figure \ref{fig:PRNU}).
Clearly, this approach relies on some important prior knowledge,
since a certain number of images taken from the source device, or the device itself, must be available.
On the other hand, it is extremely powerful, as it can detect equally well all attacks, irrespective of their nature.
The key problem is the single-image estimation at test time,
since the PRNU pattern is a weak signal, easily overwhelmed by imperfectly removed image content.
To reduce false alarms, in \cite{Chen2007} a predictor is designed to adapt the decision threshold to the local image statistics,
while in \cite{Chen2008} disturbing non-unique artifacts are detected and removed.
In \cite{Chierchia2014} the strong spatial dependencies are modeled through a Markov Random Field so as to make joint rather than isolated decisions.
Further variations rely on the use of guided filtering \cite{Chierchia2014guided}, discriminative random fields \cite{Chakraborty2017} and multiscale analysis \cite{Korus2017}.

It is worth noting that this approach can be also extended to blind scenarios,
where no prior information about the camera is known
provided a suitable clustering procedure identifies the images which share the same PRNU \cite{Cozzolino2014b, Cozzolino2017PRNU}.

An alternative to using PRNU is to base the analysis on camera {\em model} local features.
Since cameras of the same model share proprietary design choices for both hardware and software,
they will leave similar marks on the acquired images.
Therefore, in \cite{Verdoliva2014} it was proposed to extract local descriptors from same-model noise residuals
to build a reference statistical model.
Then, at test time, the same descriptors are extracted in sliding-window modality from the target noise residual and compared with the reference.
Strong deviations from the reference statistics suggest the presence of an attack.
With respect to PRNU-based analysis, this approach cannot discriminate devices, but only models.
On the other hand, model-related artifacts are much stronger than device-related PRNU, and provide more reliable information.

\begin{figure*}[t!]
	\centering
	\begin{tabular}{cc}
		\includegraphics[width=1.05\linewidth, trim=20 100 0 10, clip]{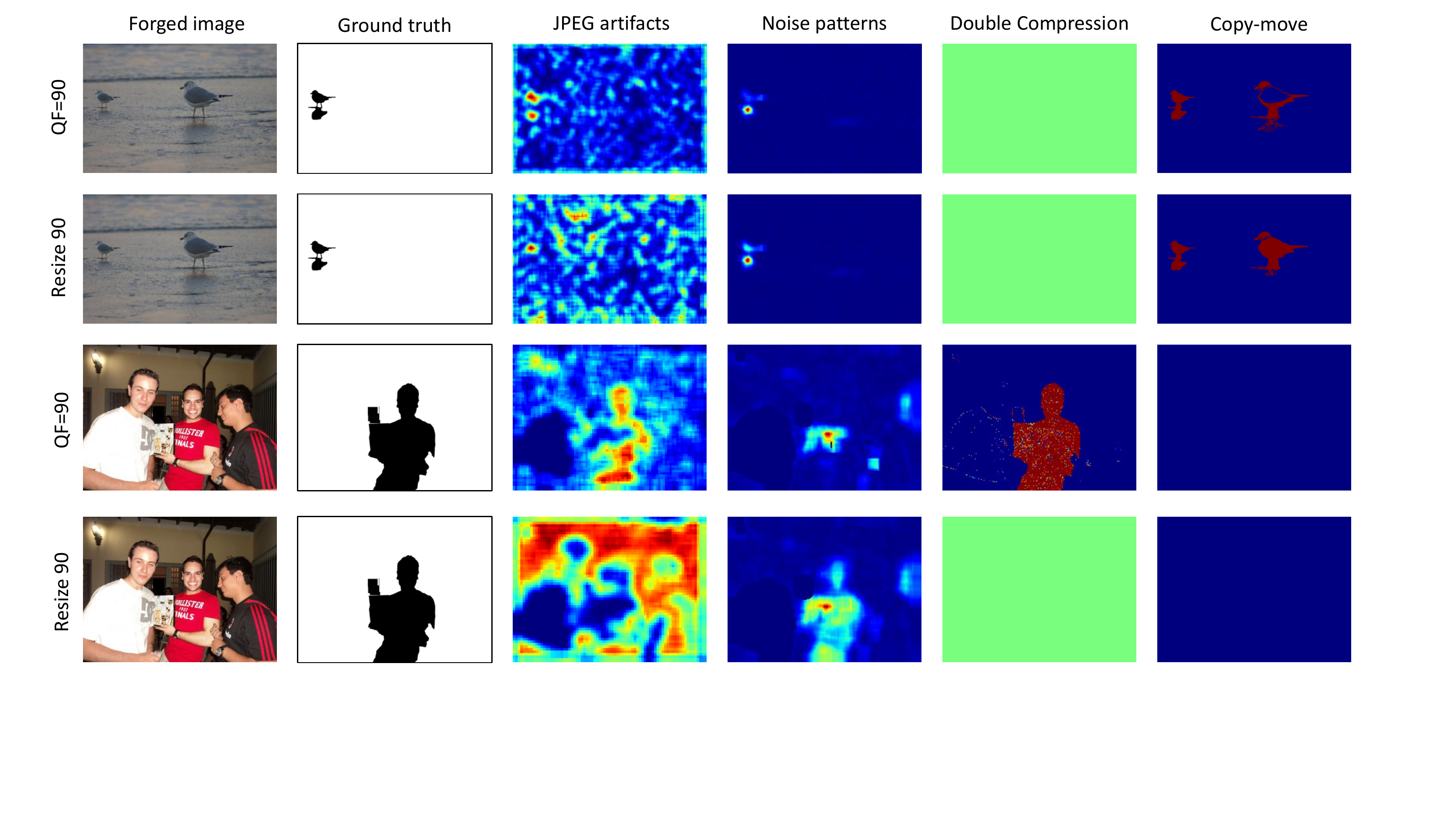}
	\end{tabular}
	\caption{Localization results of the same blind methods of Figure \ref{fig:blind_uncompressed} after compression (QF=90) or resizing (scale=90\%).
    Model-dependent methods, based on JPEG or double quantization artifacts, suffer a much stronger performance impairment than methods based on noise pattern anomalies.}
	\label{fig:blind_compressed}
\end{figure*}

\subsection{Supervised methods with handcrafted features}

These methods are based on machine learning.
Suitable features are first defined which help discriminating between pristine and manipulated images,
and then a classifier is trained on a large number of examples of both types.
It is worth underlining that features are {\em hand-crafted} by the forensic analyst,
based on a deep understanding of the target manipulations.

Some features have been devised to detect specific artifacts,
especially those generated by double JPEG compression \cite{He2006, Chen2008, Jiang2018} 
or related to the camera response function (CRF) \cite{Lin2005, Ng2009, Hsu2010}.
However, more precious are the {\em universal} features, based on suitable image statistics,
which allow detecting many types of manipulation.
Therefore, good statistical models for natural images may help selecting features that guarantee the highest discriminative power.
As already observed, to highlight statistical anomalies caused by manipulations,
one should first remove the high-level image content, to be regarded as noise \cite{Bayram2006}.
Therefore, the most effective features are typically extracted from noise residuals,
either in the spatial \cite{Zhao2013, Li2018} or in a transform domain \cite{Lyu2005, He2012}.
The pioneering work of Farid and Lyu \cite{Farid2003}, back in 2003, proved the potential of features based on high-order image statistics.
These features capture subtle variations in the image micro-textures
and prove effective in many application fields, such as computer graphics, biometrics, and steganalysis.
Therefore, it is not by chance that the most popular such features, known as rich models \cite{Fridrich2012},
have been originally proposed for steganalysis and later applied with success in forensics.
After passing the image through a set of high-pass filters, each one able to capture slightly different artifacts,
the features are formed based on co-occurrence of selected neighbors.
Then, an ensemble classifier is built.
In 2013, two methods \cite{Cozzolino2014a, Cozzolino2014b} based on the fusion of these features
and other forensic tools \cite{Chierchia2014, Cozzolino2015efficient}
ranked first in both the detection and localization phases of the first IEEE IFS-TC Image Forensics Challenge.

\subsection{Discussion}

A major appeal of blind methods is that they do not require further data besides the image/video under test.
However, methods based on very specific details depend heavily on their statistical model, and mostly fail when the hypotheses do not hold.
With reference to Figure \ref{fig:blind_uncompressed}, for example,
methods based on JPEG artifacts localize correctly both a copy-move and a splicing.
However, if the image is slightly compressed (QF=90) or resized (scale 90\%), as usual on social networks,
their performance drop dramatically, as shown in Figure \ref{fig:blind_compressed}.
Copy-move detectors, instead, are more reliable, even in the presence of post-processing, but can only detect cloning and some types of inpainting.
On the contrary,
methods based on noise patterns are quite general, and robust to post-processing,
as they often do not depend on explicit statistical models but look for anomalies in the noise residual.
Moreover, to improve reliability, they can be used in a supervised modality, as shown in Figure \ref{fig:Goldberg}.
The analyst can select a suspect region of interest for testing, using the rest of the image as an intrinsic model of pristine data.

As for machine learning-based methods, they can achieve very high detection results:
in the 2013 challenge the accuracy was around 94\% \cite{Cozzolino2014a}.
However, performance depends heavily on the alignment between training set and test data.
It is very high when training and test sets share the same cameras, same types of manipulation, same processing pipeline,
like when a single dataset is split in training and test or cross-validation is used.
As soon as unrelated datasets are used, the performance drops, sometimes close to 50\%, that is, random guess.
Lack of robustness limits the applicability of learning based approaches to very specific scenarios. 

%% file: deep_learning.tex
\section{Deep learning-based approaches}

Recently, much attention has been devoted to deep learning-based methods,
where features can be directly learnt from the data.
Deep learning has proven successful for many computer vision applications, largely advancing the state-of-the-art.
Is the same happening in multimedia forensics?
How are deep learning ideas and architectures adapted to address the specific challenges of this field?
This Section will describe deep learning-based methods proposed for the detection of generic manipulations.
Then, next Section reviews methods for the detection of GAN-generated images and video deepfakes.

\begin{figure}[t!]
	\centering
	\begin{tabular}{cc}
		\includegraphics[width=2.05\linewidth, trim=50 170 0 50, clip]{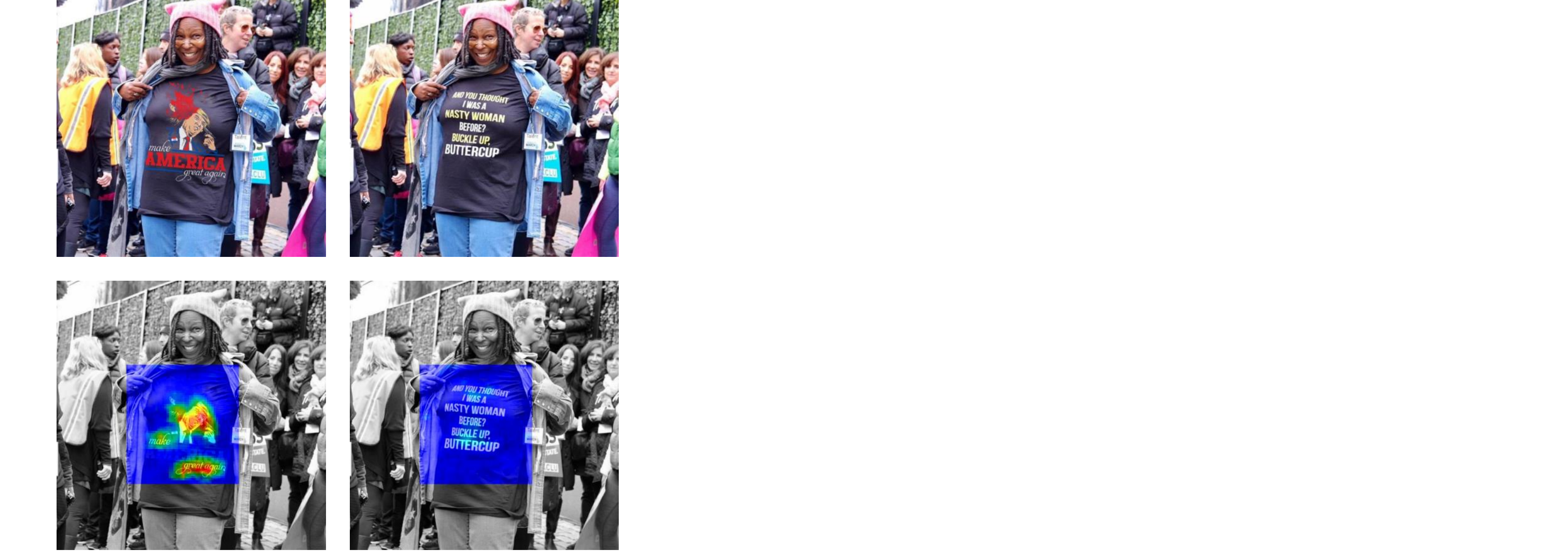}
	\end{tabular}
	\caption{Using single-asset anomaly-based methods in supervised modality.
		If only a well-defined region of interest (RoI) might have been manipulated, as with these two versions of Whoopy Goldberg,
		one can restrict the analysis to the RoI, and use the rest of the image as intrinsic model of pristine data. %\cite{Cozzolino2015}.
		A strong anomaly appears in the left version, not in the right one.}
	\label{fig:Goldberg}
\end{figure}

\subsection{Supervised CNNs looking at specific clues}

Some papers propose CNN architectures to detect specific artifacts generated by the editing process.
Double JPEG compression, as seen in the previous Section, provides strong clues for authenticity verification.
So, \cite{Wang2016} uses the histograms of DCT coefficients as input to the CNN,
while \cite{Park2018} extracts block-wise histogram-related statistical features, so as to enable also localization.
To better exploit the peculiarities of forensics as well as the learning ability of CNNs,
the approach proposed in \cite{Barni2017} works on noise residuals rather than on image pixels and uses the first layers of the CNN to extract histogram-related features.
Good results are achieved also when test images are compressed with quality factors (QFs) never seen in training and when conventional methods fail (second QF larger than the first one).
To improve performance, \cite{Amerini2017} uses a multi-domain approach relying on both spatial domain and frequency domain inputs.

Double compression detection has been also extended to H.264 video analysis in \cite{Nam2019},
where a two-stream neural network is proposed to analyze separately intra-coded frames and predictive frames.
A method specifically devised to detect sequence duplication in videos is proposed in \cite{Long2019}.
First coarse-level matches between candidate clones are identified,
and then a Siamese network based on the ResNet architecture \cite{He2016} identifies fine-level correspondences.

Other useful editing inconsistencies arise with composition.
When material from different sources is spliced together, artifacts can arise at the boundaries.
In \cite{Salloum2018} a multi-task fully convolutional network is devised,
which includes a specific branch to detect boundaries between inserted regions and background
and another branch for the surface of the manipulation,
while in \cite{Bi2019} a segmentation network based on U-Net is proposed.

Deep learning methods have been also applied to detect copy-move forgeries.
A first solution has been proposed in \cite{Wu2018copy}, 
where an end-to-end CNNs based approach implements the three main steps of a classic copy-move solution, 
i.e. feature extraction, matching and post-processing to reduce false alarms.
This helps to jointly optimize all the modules and gives as output the predicted forgery localization maps.
In \cite{Zhong2019} a different architecture is devised, which includes a multiscale feature analysis and a hierarchical feature matching, which seems to better adapt to different scenarios.
Overall, deep learning based methods perform better on low resolution images with respect to conventional approaches, where probably parameters have been adapted to high resolution images.
More interestingly it is the analysis carried out in \cite{Wu2018buster, Barni2019copy},
where the problem of source-target disambiguation is faced. 
In fact, copy-move methods typically generate a map of the original object and its clone,
and do not establish which is the forgerd region.
To this end, both in \cite{Wu2018buster} and in \cite{Barni2019copy},
a two-branch architecture is proposed followed by a fusion module,
but the manipulation detection module in \cite{Barni2019copy} focuses
on the presence of interpolation artifacts and boundary inconsistencies.

Finally, in \cite{Wang2019} a CNN based solution is devised to detect artifacts 
introduced by a specific Photoshop tool,
Face-Aware Liquify, which performs image warping of human faces.
The model is trained on fake images automatically generated by the very same tool.
To increase robustness, data augmentation includes resizing (bicubic and bilinear), 
JPEG compression, and various types of histogram editing.

\subsection{Generic supervised CNNs}

Generic CNN detectors do not look for specific types of manipulations and artifacts.
Of course, training such nets is very challenging, due to the variety of possible attacks and data histories.

A first group of methods take inspiration from the handcrafted rich models features proposed in \cite{Fridrich2012}
and used with success in image forensics.
In \cite{Rao2016} and in \cite{Bayar2016}, inspired by a similar solution used in \cite{Qian2015} for steganalysis,
a CNN is proposed with a constrained first layer that performs high-pass filtering of the image,
that suppress the scene content and allow to work on residuals.
In \cite{Rao2016}, the fixed rich-model filters are adopted,
and the network is used only for feature extraction, followed by a SVM in charge of making the global decision.
In \cite{Bayar2016}, instead, constrained filter weights are learnt during training, and the CNN is used also for classification.
A slightly different perspective is considered in \cite{Cozzolino2017}
where a CNN architecture is built to exactly replicate the behaviour of the original rich-model classifier \cite{Fridrich2012}.
Then, minor architectural relaxations allow the network to be fine-tuned on domain-specific data and further improved.
Therefore, a compact network is obtained which can be trained also on scarce data.
All the methods described above allow for the detection of small patches. Localization is then possible by using a sliding window analysis on the whole image.

All these solutions rely mostly on low-level features, assuming that high-level features do not help detecting possible manipulations.
However, imperfect image editing, like badly spliced material, may well leave traces, such as strong contrast or unnatural tampered boundaries.
Hence, a two-stream network is proposed in \cite{Zhou2017} and extended in \cite {Zhou2018}.
On the first path, rich model filters are used again to extract low-level features,
while a second path relies on the RGB data to look for high-level traces.

Fixed high-pass filters in the first layer have been used also more recently in \cite{Li2019},
and in combination with standard filters in \cite{Wu2019}.
However in both these papers a fully connected network is proposed in order to obtain as output the binary localization map, that hence accounts also for detection at image level.
In \cite{Li2019} pixel-wise predictions are obtained by applying transpose convolutions.
It is also observed that typically the area of a manipulated region is much smaller than the untouched pixels,
and a focal loss is adopted to face this class imbalance, that assigns a modulating factor to the cross entropy term. This paper mostly focuses on manipulations created using deep learning based inpainting methods, while the objective of the work proposed in \cite{Wu2019} is to detect every type of possible local manipulations.
To this end, the training procedure is built so as to classify 385 image manipulation types,
and to detect features related to local anomalies. 
Another generic solution to detect and localize image manipulations has been proposed in \cite{Bappy2019}.
Resampling features are used to capture inconsistencies, long short-term memory (LSTM) cells highlight transitions between pristine and forged blocks in the frequency domain, 
and finally an encoder-decoder network segments the manipulation.
Localization is indeed carried out in \cite{Zhou2019adv} by first introducing a
process to generate forged (harder) examples during training 
in order to generalize across a large variety of possible manipulations.
Then, a segmentation and refinement network is used so that the algorithm is forced to look 
at boundary artifacts.

In \cite{Rahmouni2017, Boroumand2018, Marra2019} the problem of image-level analysis for forgery detection
is specifically addressed.
Some methods \cite{Rahmouni2017, Boroumand2018} train the CNN to extract compact features at the patch level,
and then perform some forms of external aggregation to make image-level decisions.
However, a patch level analysis does not allow to take into account at the same time both local (textural analyses) and global (contextual analyses) information.
This latter requirement is not easily met, because CNNs accept in input patches that are much smaller than the whole image needed for contextual analysis.
In computer vision, this problem is solved by resizing the image,
but this process destroys the fine-grain structure of the image and hides important traces of manipulation \cite{Boroumand2018, Marra2019}.
In \cite{Marra2019} a gradient checkpointing is used to allow end-to-end training of both aggregation and feature extraction,
allowing for their joint optimization, without any resizing.
This helps analyzing the whole image through textural-sensitive features and highlight anomalies that would not appear at the patch level.

\subsection{One-class training}

Assembling a training set representative of all possible manipulations can be a prohibitive task.
Hence, an alternative is to resort to a one-class approach and look for anomalies with respect to an intrinsic model of pristine data.
Indeed, any manipulation is by definition an anomaly, and should be detectable as such.
These methods possess then the desirable property to detect any type of manipulations.

In \cite{Cozzolino2016} a single-asset (blind) one-class method has been proposed.
Expressive features are extracted from the noise residual through an autoencoder, and iterative feature labeling singles out two classes.
The largest class defines the pristine model.
Then, various criteria can be used to decide on whether the data of the second class are also pristine or else manipulated.
In \cite{Davino2017}, the approach has been extended to videos by including a LSTM recurrent network to account for temporal dependencies.
In \cite{Yarlagadda2018}, instead, GANs are used to learn features typical of pristine images,
followed by a one-class SVM trained on them to determine their distribution, and eventually detect tampering of satellite images.

Several papers leverage the strong connection existing between source identification and splicing detection and localization.
Indeed, in the presence of a splicing, the fact that different image parts are acquired by different camera models provides powerful forensic clues.
In \cite{Bondi2017a}, a CNN is used to extract camera-model features from image patches, followed by clustering to detect anomalies.
A similar approach is followed in \cite{Mayer2018}.
First, a constrained network is used to extract high-level camera-model features, then, another network is trained to learn the similarity between pairs of such features.
This work has been recently extended in \cite{Mayer2019} introducing a graph-based representation
that better captures the forensic relationships among all image patches within an image. 
A Siamese network is also trained in \cite{Huh2018} to decide whether two image patches have similar metadata.
Once trained on pristine images with EXIF header, the network can be used on any image without further supervision.
In \cite{Cozzolino2020,Cozzolino2018a,Cozzolino2019} these concepts are exploited to extract a camera-model fingerprint, called noiseprint,
similar to a PRNU-based device fingerprint.
A denoiser CNN is trained in Siamese modality to tell apart similar (same camera model and same position) from different couples of patches.
Once trained, the network is able to extract the image-size noiseprint, where artifacts related to camera model are emphasized.
In \cite{Cozzolino2020} it is shown that noiseprints, thanks to their spatial sensitivity, can be used to detect splicing as well as several other manipulations, while in \cite{Cozzolino2019} the approach is extended to video forensics.

%% file: deepfakes.tex
\section{Deepfake detection}

Human faces are by far the most expressive and emotionally-charged pieces of information that circulate on the web.
The face is the main biometric trait of a person, a universal ID card, 
and a vehicle itself of non-verbal but powerful messages.
Therefore,
the appearance of artificial intelligence-powered tools that
generate realistic faces of persons that do not exist, 
or modify in a credible way the attributes of faces in videos, has raised great alarm.

However, computer generated faces already existed before the deep learning era.
Research on distinguishing real from computer generated faces has been going on for years and represents a precious starting point.
Indeed, fakes generated with CGI and deep learning have much in common,
since they both lack the characteristic features that are typical of images and videos of human faces acquired by real cameras.
In \cite{Nguyen2012} face asymmetry is proposed as a discriminative feature to tell apart computer generated from real images of human faces.
Then, in \cite{Nguyen2015} the focus shifts to videos,
and detection relies on the spatial-temporal deformations of a 3D model that fits the face.
In particular, natural faces follow more complex and various geometric deformations then synthetic ones, and cause higher perturbations of the 3D model.
Also, natural faces belong to living persons.
So, the method proposed in \cite{Conotter2014} relies on the small variations in the appearance of the face due to the periodic blood flow caused by heart beating.

The following subsections review the work that has been devoted explicitly to detect local manipulations to images or videos or fully sinthetic media created using deep learning strategies.
First, the methods based on handcrafted features will be described, then those relying themselves on deep learning will be analyzed.

\begin{figure}[t!]
	\centering
	\includegraphics[width=0.195\linewidth]{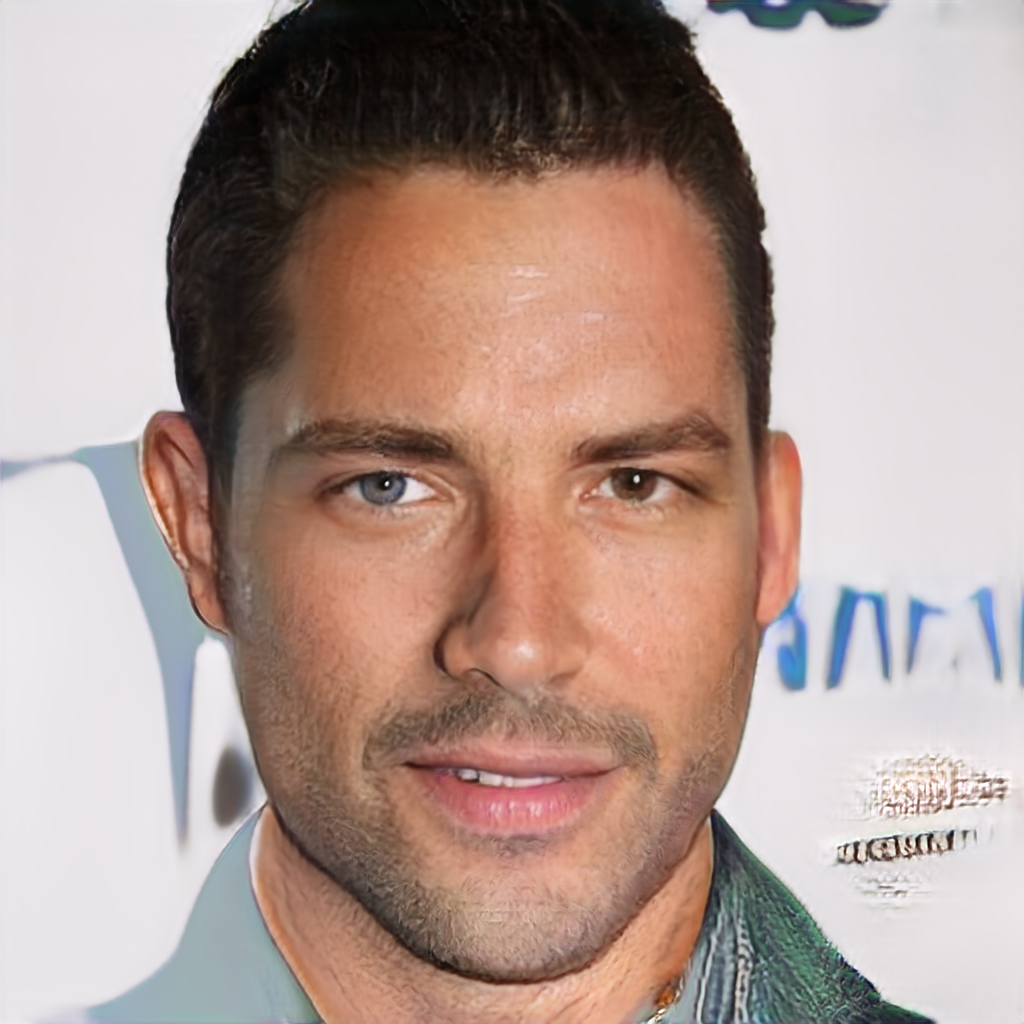}
	\includegraphics[height=0.195\linewidth]{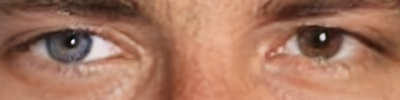} \\[0.2em]
	\includegraphics[height=0.237\linewidth]{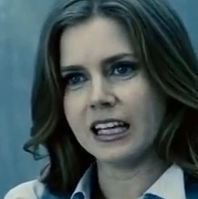}
	\includegraphics[height=0.237\linewidth]{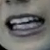} 
	\includegraphics[height=0.237\linewidth]{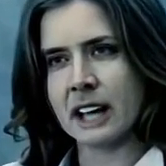}
	\includegraphics[height=0.237\linewidth]{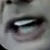} 	
	\caption{Today's deepfakes sometimes exhibit some obvious asymmetries, such as eyes of different colors (top) or badly modeled teeth (bottom). However, such artifacts will likely disappear in the future.}
	\label{fig:artifacts}
\end{figure}

\subsection{Methods based on handcrafted features}

A rather general approach is to look for high-level visual artifacts in the face.
Methods following this approach try to highlight specific failures in the generation process which does not reproduce perfectly all the details of a real face.
For example, in GAN-generated faces a mismatch may occur between the color of the left and right eye,
as well as other forms of asymmetry, like a earring only on one side, or ears with markedly different characteristics.
Deepfakes, instead, often present unconvincing specular reflections in the eyes, either missing or represented as white blobs,
or roughly modeled teeth, which appear as a single white blob (see Figure \ref{fig:artifacts}).
All these artifacts are exploited in \cite{Matern2019}, where simple features are built in order to capture them.
\cite{Li2018ictu} relies on eye blinking, which has a specific frequency and duration in humans, which is not replicated in deepfake videos.
A solution based on a long-term recurrent network, working only on eye sequences, is designed to catch such temporal inconsistencies.
In \cite{Ciftci2019} and in \cite{Fernandes2019} deepfakes are revealed by the lack of variations induced by heart beating,
an idea already exploited in \cite{Conotter2014} for computer generated faces.
However, in \cite{Ciftci2019} the coherence of these biological signals is considered both spatially and along the temporal direction.

Other detection methods rely on face warping artifacts \cite{Li2019warping}, face landmark locations  \cite{Yang2019landmark} or head pose inconsistencies \cite{Yang2019head}.
In \cite{Li2019warping} the approach exploits the fact that current deepfakes generation methods
are able only to generate limited resolution images, 
that need to be further warped to match the original face in the source video.
However, warping leaves peculiar traces that can be detected using a CNN that works on the face region and its surrounding areas.
Instead, the observation made in \cite{Yang2019landmark} is that 
GAN-based face synthesis algorithms are able to generate a face with high level of realism
and with many details, but lack an explicit
constraint over the locations of these parts in a face.
Hence, the locations of the facial landmark points, like the tips of the eyes, nose and the
mouth, can be used as the discriminative features for verifying the authenticity of GAN images.
This same problem is also present in deepfake videos and can be revealed 
by means of 3D head pose estimation \cite{Yang2019head}.

Other common artifacts of GAN-generated images are related to how color is synthesized.
In fact, the generator is constrained so as to limit the occurrence of saturated and under-exposed pixels \cite{McCloskey2019}, not infrequent in real images.
Other disparities in color components are exploited in \cite{Li2018color},
in fact deep networks generate images in the RGB color space without any type of constraint on color correlations,
and artifacts arise if looking at features in other spaces 
such as HSV and YCbCr, especially in the chrominance
components. 

A clear advantage of all these methods is that visual artifacts are not affected by resizing and compression.
On the other hand, fake media that can be recognized also by human viewers represent less of a menace.
Moreover, with the current pace of technology,
it is very likely that next-generation deepfakes will overcome such imperfections and synthesize visually perfect fakes.

A different approach is followed in \cite{Agarwal2018}.
The idea is to protect individuals by acquiring some peculiar soft traits that characterize them and are very difficult to reproduce for a generator.
In particular, it is observed that facial expressions and head movements are strongly correlated,
and changing the former without modifying the latter may expose a manipulation.
On the down side, to apply this approach, a large and diverse collection of videos in a wide range of contexts must be available for all individuals of interest.

\subsection{Methods based on deep learning}

With reference to GAN images, a first investigation has been carried out in \cite{Marra2018},
where several CNN architectures have been tested in a supervised setting to discriminate GAN images from real ones.
Several solutions appear to be very effective,
but the performance decreases significantly when training and test mismatch, or when data are compressed using the pipeline typical of social networks.
In \cite{Marra2019_DoGAN}, as a preliminary step for forensic analyses,
it is shown that each specific GAN architecture is characterized by its own artificial fingerprint, present in all generated images,
much like a real camera is characterized by its PRNU pattern.
A similar goal is pursued in \cite{Yu2019} through a suitably trained CNN.
This last work also shows that these fingerprints persist across different image frequencies and patches, and are not biased by GAN artifacts.
In \cite{Zhang2019}, instead, a GAN simulator is proposed to reproduce common GAN-image artifacts, which manifest as spectral peaks in the Fourier domain.
Then, a classifier is trained, which takes the spectrum as input.
Another work \cite{Albright2019} tries to attribute a test image to a specific generator in a white-box scenario.
The basic idea is to invert the generation process.
Once the original latent vector is recovered, it is fed again to the network, and the output is compared with the test image. A similar idea is also suggested in \cite{Karras2019dec}, where 
using projection-based methods the authors show that it is possible to detect 
that an image was synthesized by a specific network, even if it is of very high quality.

\begin{figure}[t!]
	\centering
	\begin{tabular}{cc}
		\includegraphics[width=1.35 \linewidth, trim=100 200 0 70, clip]{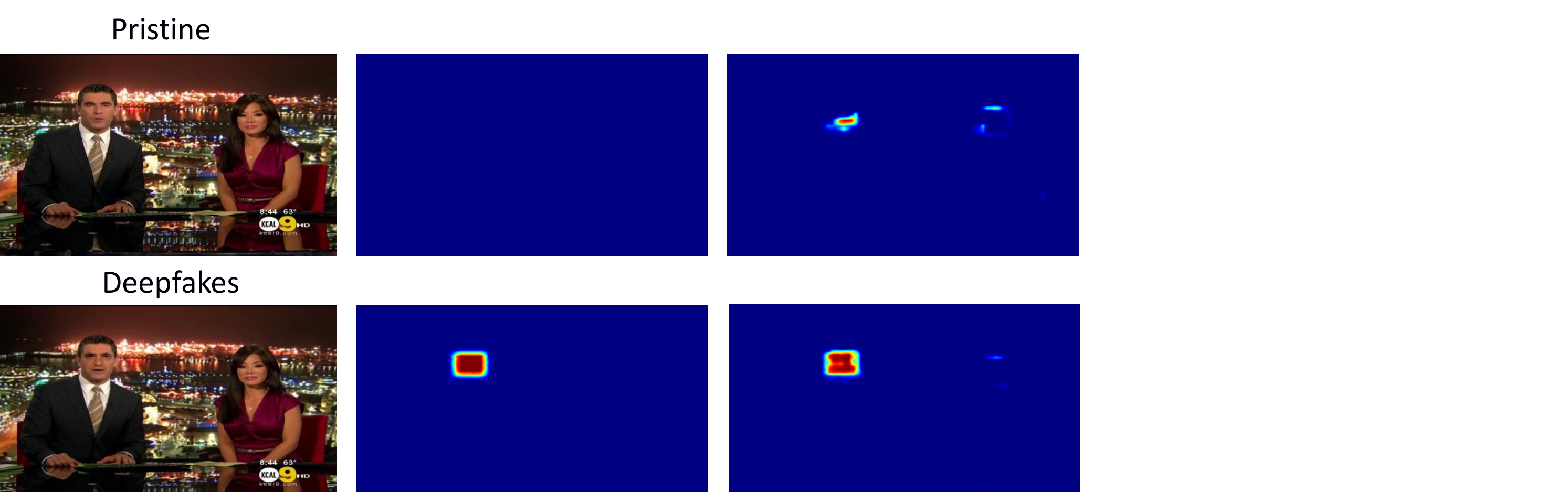}
	\end{tabular}
	\caption{Localization results provided by a deep network on a pristine video (top) and a deepfake (bottom).
    Excellent results are obtained with high-quality videos, while artifacts appear if the videos are compressed at low bit-rates.}
	\label{fig:faceforensics}
\end{figure}

Switching to solutions for deepfake detection in videos,
in \cite{Afchar2018} two simple architectures are proposed with a small number of levels and parameters
that exploit mesoscopic features.
A first solution (Meso-4) has four layers of convolutions and pooling and is then followed by
a dense network with one hidden layer. The second solution (MesoInception-4) instead is based on a variant of the inception module that includes dilated convolutions.
However, experiments carried out in \cite{Roessler2019} clearly show that, in a supervised setting,
very deep general-purpose networks \cite{Szegedy2016, Huang2017, Chollet2017},
outperform forensics-oriented shallow CNNs, as well as methods based on handcrafted features,
especially in the presence of the strong compression typical of video codecs.
For the detection task these methods use specific strategies to extract the faces from the frames,
as input to the network. However, also localization strategies can be devised on the whole frame.
In Fig.\ref{fig:faceforensics} the pixel-level localization results of a deep network are presented 
using the approach described in \cite{Roessler2018}, however this analysis can be accomplished in different ways \cite{Li2019zoom}.
Training on uncompressed data ensures perfect results on high-quality videos, while artifacts appear when the test video is strongly compressed.
Further improvements can be obtained by including an attention mechanism, as done in \cite{Stehouwer2019}.
Promising results are also obtained in \cite{Nguyen2019} by using capsule-network architectures which require fewer parameters to train than very deep networks.
It should be noted that these methods operate on still frames.
Therefore, to boost performance, the scores obtained at frame level can be aggregated \cite{Afchar2018},
or else an ensembling approach can be applied, using different color spaces as input \cite{He2019}.

Clearly, better results can be expected by strategies that take explicitly into account the temporal direction.
In fact, even if current generation methods are very effective,
they perform face manipulation on a frame-by-frame basis and hence may incorrectly follow the face movements.
Several methods have been proposed in the literature to exploit this point.
In \cite{Guera2018, Sohrawardi2019} 
a convolutional Long Short Term Memory (LSTM) network is used to exploit such dependencies
and improve upon single-frame analysis.
In \cite{Sabir2019}, instead, a solution based on recurrent convolutional models has been proposed.
Features are extracted at multiple levels and processed in individual recurrent networks,
in order to exploit micro, meso and macroscopic features for manipulation detection.
A significant improvement can be achieved even by using only 5 frames. 
In \cite{Amerini2019} the optical flow field is estimated to exploit discrepancies in motion across frames, 
but only some preliminary results are presented.

Despite these interesting results,
it must be underlined that most of these detectors tend to overfit the training set, and perform badly on new data \cite{Khodabakhsh2018}.
This is becoming a major issue in multimedia forensics,
considering the fast growing number of deep learning-based manipulation methods.
Therefore, new proposals should ensure good generalization ability,
which calls for better validation of detectors, with multiple datasets and different types of manipulations.
Focus on this problem is first found in \cite{Cozzolino2018}, 
where an autoencoder-based architecture is proposed which adapts to new manipulations with just a few examples.
This same approach has been followed in \cite{Nguyen2019a} by using a deeper network and including a segmentation task. An autoencoder is also used in \cite{Du2019}, where
a pixel-wise mask is used to enforce the
model to learn intrinsic representation from the forgery region, so as to avoid to detect artifacts
present in the training set.
In \cite{Marra2019incremental} it is proposed a method based on incremental learning
to reduce the burden and the risks of re-training networks on larger and larger datasets as new forms of manipulation appear.
A different perspective is followed in \cite{Xuan2019},
where a pre-processing step is introduced in order to 
reduce low level artifacts of GAN images and force the discriminator to learn more general forensic features.
Instead, in \cite{Wang2019dec} a careful pre- and post-processing and data augmentation are applied to 
improve transferability. The work shows that CNN-generated images
share some common flaws that allow one to trace their origin even on unseen architectures, datasets and training methods.
The main idea is to make a very large augmentation in the training step by means of 
several and different post-processing operations, like blurring and compression and combinations of them, 
even if they are not performed at test time.

Turning to videos, recently some interesting solutions have been considered to improve generalization. 
In \cite{Fernando2019} this is achieved using hierarchical neural memory networks.
Beyond exploiting long-term dependencies, the method includes an attention mechanism and 
an adversarial training strategy. This helps to increase robustness to compression and to 
transfer learning to better deal with unseen manipulations.
A completely different perspective is followed in \cite{Li2019dec}.
In this work the focus is on the boundary of forged faces. 
In fact the observation is that they all share a common blending operation. 
Hence the generalization ability of the network increases,
since it is not based on the artifacts of a specific face manipulation method.
In \cite{Cozzolino2019} generalization is gained by training the method only on pristine videos
and by extracting the camera fingerprint information (noiseprint) gathered from multiple frames.
This can significantly improve the detection results on various types of face manipulations,
even if the network never saw such forgeries in training.

%% file: considerations.tex
\section{Deep learning in multimedia forensics: considerations}

The review of previous Sections
testifies of an exponential growth in the number of papers proposing deep learning methods for multimedia forensic problems.
But, where are we now? Are forensic analysts winning or losing their ``war''? Which lessons have we learned?
Amidst all these proposals,
it is not easy to extract truly innovative ideas, solid scientific trends, effective and robust solutions.

For data-driven methods, lacking theoretical models,
experimental results take a fundamental role, and hence experimental protocols are extremely important.
Let us consider supervised methods, for the time being,
with perfectly aligned training and test sets, that is, disjoint training and test samples drawn from the very same dataset/distribution.
Experimental evidence shows that, in this setting, deep learning methods work extremely well.
In ideal conditions this is not so important, because then simpler approaches work just equally well.
In challenging conditions, however, when forensic traces are weak,
deep learning, especially very deep architectures, can provide a large performance gain with respect to conventional methods.
In Table \ref{tab:robustness}, to compare approaches based on handcrafted features, deep networks, and very deep networks, 
accuracy results on DeepFakes videos are reported from \cite{Roessler2019}.
In the presence of strong compression, there is a gap of about 15\% between machine learning and deep learning, and 15\% more using a very deep network.
This can make a big difference in practical applications,
since many standard processing steps tend to weaken forensic traces,
as when video are routinely compressed and resized as soon as they are uploaded on a social network.
This is certainly a remarkable achievement of deep learning methods.

However,
a validation protocol which considers only perfect alignment is intrinsically weak, and falls short of its goal.
Indeed, perfect alignment is a very favorable setting,
which is easily obtained in simulations and rarely observed in real-world operations.
Main causes of misalignment are:
{\it  i)} the target media asset has been generated or manipulated in ways that were never seen in the training set;
{\it ii)} its processing history is not covered by training set samples.
Both situations are extremely common.
New forms of manipulations are invented by the day, and cannot be represented in the training set when they first appear.
Moreover, images and videos can undergo a long sequence of transformations \cite{Bianchi2015} and accounting for all of them is not reasonable.
Working well with aligned training and test sets, possibly carved from the same dataset, is not very informative.
Therefore, it is important to adopt stronger validation protocols,
including experiments where training and test set are unrelated and account for realistic and challenging conditions,
such as compression, multiple manipulations, unseen forgeries.
In the absence of such strong validation, the results of the supervised deep learning methods do not yet appear completely convincing.

\begin{table}[t!]
	{\footnotesize
		\centering
		\caption{Results of CNN-based methods on Deepfakes}
		\label{tab:robustness}
		\begin{tabular}{l||c|c|c|} \cline{2-4}
			\ru                          & \multicolumn{3}{c|}{Accuracy} \\ \cline{2-4}
			\ru                          & Uncompressed & High Quality & Low Quality \\ \hline\hline
			\ru  Handcrafted features    & 99.03\% & 77.12\%  & 65.58\% \\ \hline
			\ru  Deep network            & 99.28\% & 90.18\%  & 80.95\% \\ \hline
			\ru  Very deep network       & 99.59\% & 98.85\%  & 94.28\% \\ \hline
		\end{tabular}
	}
\end{table}

One-class methods seem to perform reasonably well also in challenging real-world conditions,
and this looks as a promising approach to deal with such complex scenarios.
Clearly, for one-class methods, no alignment problem exists.
On the other hand, the fundamental question of what can be labelled as ``pristine'' remains open.
Many operations are not malicious and should not be detected, yet they change the statistics of an image with respect to what is output by a camera.
For example, one can apply histogram equalization to improve the appearance of a photo, but that is not a forgery.
Even resizing has a different meaning based on the context: 
it is a sign of a manipulation when used to change the dimension of an object to perform splicing, while it is an innocent operation when used to save space.
For these reasons, looking for local anomalies seems to be a good direction, which overcomes ambiguities.
In general, the training phase is crucial to make a network work in the correct and desired way,
and there is a strong need of large datasets that try to cover many different situations.

%% file: datasets.tex
\section{Datasets}

For learning based approaches, it is of paramount importance to have good data for training.
Moreover, to assess the performance of new proposals, it is important to compare results on multiple datasets with different features.
The research community has made considerable efforts through the years to release a number of datasets
with image and video manipulations.
However, not all of them possess the right properties to support the development of learning-based methods.
Many such methods, in fact, split a single dataset in training, validation, and test set,
a practice that may easily induce some forms of polarization or over-fitting if the dataset is not built with care.
In this Section, the most widespread datasets are described, and their features briefly discussed.

\subsection{Images}

Table \ref{tab:datasets_images} reports a list of datasets with manipulated images.
Some of them are rather old and necessarily outdated, and some even present important flaws.
It is surprising to see recent papers relying on unsuited datasets and presenting them as challenging testbeds.

\begin{figure}[t!]
	\centering
	\begin{tabular}{cc}
		\includegraphics[width=1.60\linewidth, trim=0 270 0 0, clip]{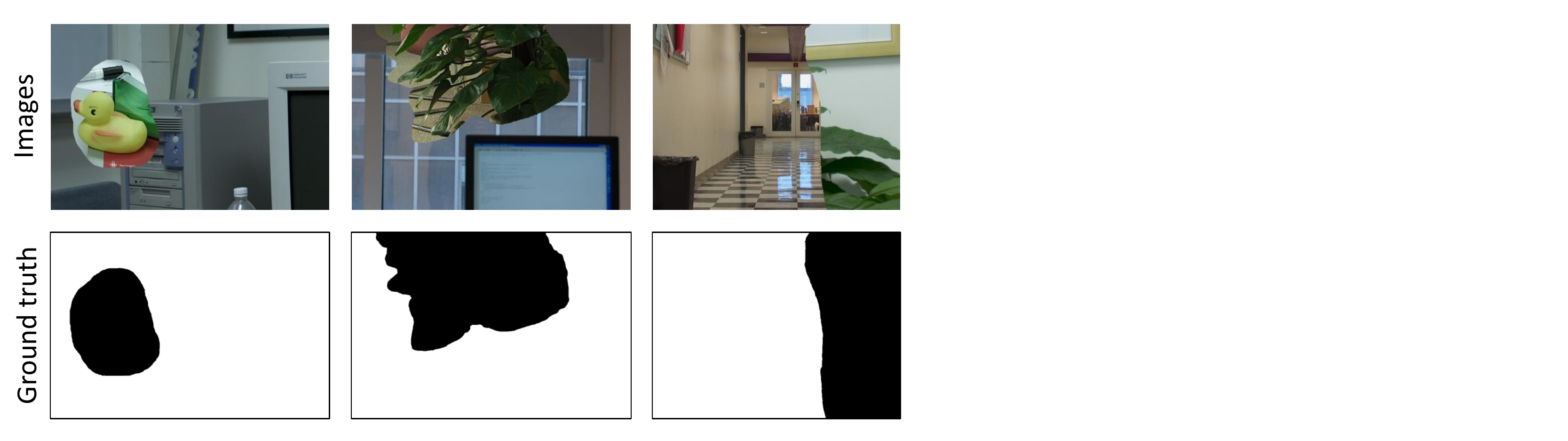}
	\end{tabular}
	\caption{Examples from the Columbia dataset. Top: images with splicing, bottom: ground truth.
		In all cases the inserted region is very large and obviously detectable.}
	\label{fig:Columbia}
\end{figure}

The Columbia (color) dataset, presented in 2006 \cite{Hsu2006}, is one of the first ones made available to the forensics community.
It comprises 180 forged images with splicing, some examples of which are shown in Figure \ref{fig:Columbia}.
Despite its merits, it presents several major problems:
{\it   i)} it is unrealistic, since the forgery is clearly visible;
{\it  ii)} the spliced region is not subject to any type of post-processing;
{\it iii)} only uncompressed images are present;
{\it  iv)} only four cameras are used to take both the host images and the spliced regions.
Moreover, it is not clear how to define the forged areas, given that both regions come from pristine data.
For these reasons, this dataset should not be used both in the training and the testing phase, nor to perform fine-tuning,
otherwise overly optimistic results will be observed.

Also widespread is the Casia dataset, proposed in 2013 \cite{Dong2013}.
In the first version (v1) splicings present sharp boundaries and are easily detectable.
The second version (v2), however, is more realistic, and inserted objects are post-processed to better fit the scene.
Nonetheless, it exhibits a strong polarization, highlighted in \cite{Cattaneo2014}.
In fact, tampered images and pristine images are JPEG compressed with different quality factors (the former at higher quality).
Therefore, a classifier trained to tell tampered and pristine images apart,
may instead learn their different processing history,
thereby working very well on test images from the same dataset, and very poorly on new unrelated images.

Another dataset with splicings is DSO-1 \cite{Carvalho2013} a subset taken from the IEEE Image Forensics Challenge
(unfortunately, the original datasets prepared for the challenge is not available anymore and the ground truths were never released by the organizers).
Here, the manipulations are carried out with great care and most of them are realistic.
Images are saved in the uncompressed PNG format, but most of them had been JPEG compressed before.
Minor problems are the fixed resolution of images,
and missing information on how the dataset was created, {\it e.g.,} how many cameras were used, which could help interpreting results.

Forgeries of various nature are present in the Realistic Tampering Dataset proposed by Korus in \cite{Korus2016a}.
The manipulated images, all uncompressed, appear indeed very realistic,
although there is only a small number of them.
The dataset includes also the PRNU patterns of the four cameras used to acquire all images, enabling the use of sensor-based methods.

The Wild Web Dataset \cite{Zampoglou2017} is a collection of real cases from the internet.
Therefore, there are no certified information on the manipulations,
but the authors made a huge effort to gather different versions of the same images and to extract meaningful ground-truths.

Many datasets have been proposed specifically for copy-move forgery detection \cite{Christlein2012, Amerini2011, Tralic2013, Cozzolino2015efficient, Wen2016},
the forgery most studied in the literature.
Some of them are designed to challenge copy-move methods by adding multiple operations on the copied object, such as rotation, resizing, change of illumination.
However, the more an object is modified, the more detectable it becomes for methods based on camera artifacts,
since the distance between its statistical properties and those of the background increases.

There is also a realistic dataset to test double JPEG compression \cite{Bianchi2012} and, recently,
a synthetic dataset of single and double JPEG compressed blocks with 1,120 quantization tables, has been released, aimed at training deep networks \cite{Park2018}.

To test algorithms in the wild, the U.S. National Institute of Standards and Technology (NIST) has released several large datasets \cite{Guan2019}.
The first one, NC2016, contains some redundancies:
each spliced photo is presented four times, JPEG compressed at low and high quality, and with and without post-processing on the splicing boundaries.
These multiple versions are meant to study how performance depends on such details.
However, this feature is never exploited in the literature.
On the contrary, several papers split the dataset in training and test carelessly,
including the same image parts in both sets, and artificially boosting the performance.
In subsequent years, NIST published three more datasets, NC2017, MFC2018, MFC2019, without the above redundancies.
These datasets are very large and present a great variety of manipulations, resolutions, formats, compression levels, acquisition devices.
Moreover, multiple manipulations are often carried out on the same image and even on the same objects.
In several cases, a separate development dataset is associated with the main one to ensure correct training of learning-based methods.
Overall, they represent a very challenging and reliable testbed for new proposals.

Recently, a very large dataset has been released in \cite{Mahfoudi2019}, called DEFACTO,
which collects over 200,000 images with realistic manipulations, including splicings, copy-moves, removals and face morphing.
Another very large dataset is the PS-Battles Dataset, where a collection of 
102,028 images is presented, each containing the original image but also a varying number
of manipulated versions \cite{Heller2018ps}. 
For what concern facial modifications
a dataset has been proposed in \cite{Zhou2017} which contains around 4,000 real and manipulated images, using two different face swapping algorithms.
A large dataset of GAN-generated images using available software \cite{Cozzolino2018, Marra2019_DoGAN} are available at \cite{FT_datasets}.

\setlength{\tabcolsep}{6pt}
\begin{table*}
	\centering
	\caption{List of datasets including generic image manipulations}
	\begin{tabular}{|l|c|c|l|c|c|c|} \hline
        \ru dataset         & ref.             & year & manipulations                & \# prist. / forged  &  image size                                   & format                  \\ \hline
        \ru Columbia gray   & \cite{Ng2004}         & 2004 & splicing (unrealistic)       &         933 / 912   &   128$\times$128                              & BMP                     \\ \hline
        \ru Columbia color  & \cite{Hsu2006}        & 2006 & splicing (unrealistic)       &         182 / 180   &   757$\times$568 - 1152$\times$768   & TIF, BMP                \\ \hline
        \ru MICC F220            & \cite{Amerini2011}     & 2011 & copy-move      &         110 / 110         &   722$\times$480 - 800$\times$600  & JPG \\ \hline
        \ru MICC F2000           & \cite{Amerini2011}     & 2011 &  copy-move         &       1,300 / 700        &   2048$\times$1536    & JPG \\ \hline
        \ru VIPP            & \cite{Bianchi2012}     & 2012 & double JPEG compres.         &         68 / 69         &   300$\times$300 - 3456$\times$5184     & JPG \\ \hline
        \ru FAU             & \cite{Christlein2012} & 2012 & copy-move                    &          48 / 48    &         2362$\times$1581  $-$ 3888$\times$2592                                      &    PNG, JPG                 \\ \hline
        \ru  Casia v1       & \cite{Dong2013}       & 2013 & splicing, copy-move          &         800 / 921   & 374$\times$256                                & JPG                     \\ \hline
        \ru  Casia v2       & \cite{Dong2013}       & 2013 & splicing, copy-move          &        7,200 / 5,123  & 320$\times$240 $-$ 800$\times$600             & JPG, BMP, TIF           \\ \hline
        \ru  DSO-1          & \cite{Carvalho2013}   & 2013 & splicing                     &         100 / 100   & 2048$\times$1536                              & PNG                     \\ \hline
        \ru  CoMoFoD          & \cite{Tralic2013}   & 2013 & copy-move                     &         260 /260   & 512$\times$512, 3000$\times$2000                         & PNG, JPG                     \\ \hline
        \ru Wild Web        &  \cite{Zampoglou2015}     & 2015 & real-world cases             &     90 / 9,657               &     72$\times$45  $-$ 3000$\times$2222                                          &       PNG, BMP, JPG, GIF                   \\ \hline
        \ru GRIP        &  \cite{Cozzolino2015efficient}     & 2015 & copy-move             &      80 / 80               &       1024$\times$768                                       &    PNG                     \\ \hline
        \ru RTD (Korus)       & \cite{Korus2016a}     & 2016 & splicing, copy-move          &         220 / 220    & 1920$\times$1080      & TIF                     \\ \hline
        \ru COVERAGE      & \cite{Wen2016}        & 2016 & copy-move          &         100 / 100    & 400$\times$486     & TIF                     \\ \hline
        \ru NC2016          &    \cite{Guan2019}                   & 2016 & splicing, copy-move, removal &      560 / 564               &      500$\times$500  $-$ 5,616$\times$3,744                                         & JPG                     \\ \hline
        \ru NC2017          &  \cite{Guan2019}                     & 2017 & various                      &         2667 / 1410 &  160$\times$120  $-$ 8000$\times$5320         & RAW, PNG, BMP, JPG           \\ \hline
        \ru FaceSwap & \cite{Zhou2017}       & 2017 & face swapping                &          1,758 / 1,927       &  450$\times$338 - 7360$\times$4912      & JPG            \\ \hline
        \ru MFC2018         &     \cite{Guan2019}                  & 2018 & various                      &        14,156 / 3,265 &  128$\times$104  $-$ 7952$\times$5304         & RAW, PNG, BMP, JPG, TIF      \\ \hline
        \ru     PS-Battles       &       \cite{Heller2018ps}                & 2018 & various                      &             11,142 / 102,028         & 130$\times$60  $-$ 10,000$\times$8558 &  PNG, JPG\\ \hline
        \ru MFC2019         &       \cite{MFC2019_dataset}                & 2019 & various                      &     10,279 / 5,750                & 160$\times$120  $-$ 2624$\times$19,680 & RAW, PNG, BMP, JPG, TIF  \\ \hline
        \ru DEFACTO         &       \cite{Mahfoudi2019}                & 2019 & various &     -- / 229,000     & 240$\times$320  $-$ 640$\times$640                              & TIF \\ \hline
        \ru GAN collection    &  \cite{Marra2019_DoGAN}  & 2019 & GAN generated                      &    356,000 / 596,000            & 256$\times$256  $-$ 1024$\times$1024  & PNG  \\ \hline
	\end{tabular}
	\label{tab:datasets_images}
\end{table*}

\subsection{Videos}

Only a few datasets are available for experiments on videos, but their number has been growing rapidly in this last year.
Creating high-quality realistic forged videos using standard editing tools is very time-consuming,
hence, only a few small datasets are available on-line featuring classic manipulations, like copy-moves and splicings
\cite{Bestagini2013, Davino2017, DAmiano2019}.
Many more, and much larger datasets include video manipulated with AI-based tools
\cite{Korshunov2018deepfakes, Khodabakhsh2018, Roessler2019, Li2019dataset, Dufour2019, Deepfake_FB, Deepfake_Kaggle, Jiang2020} (Table \ref{tab:datasets_videos}).

\setlength{\tabcolsep}{6pt}
\begin{table*}
	\centering
	\caption{List of datasets including video manipulations}
	\begin{tabular}{|l|c|c|l|c|c|c|} \hline
		\ru dataset          & ref.                    & year & manipulations               & \# prist. / forged  &  frame size                       & format          \\ \hline
		\ru DF-TIMIT         & \cite{Korshunov2018deepfakes} & 2018 & deepfake        &        -- / 620     & 64$\times$64 $-$ 128$\times$128   & JPG            \\ \hline
		\ru FFW              & \cite{Khodabakhsh2018}        & 2018 & splicing, CGI, deepfake  &          -- / 150     & 480p, 720p, 1080p                                   &    H.264, YouTube             \\ \hline
		\ru FVC-2018     & \cite{Papadopoulou2018}       & 2018 & real-world cases      &      2,458 / 3,957    & various                           & various         \\ \hline
		\ru FaceForensics++  & \cite{Roessler2019}           & 2019 & deepfake, CG-manipulations &      1,000 / 4,000    & 480p, 720p, 1080p                 & H.264, CRF=0, 23, 40 \\ \hline
		\ru  DDD &   \cite{Dufour2019}          & 2019 & deepfake &      363 / 3,068    & 1080p                 & H.264, CRF=0, 23, 40 \\ \hline
		\ru Celeb-DF         & \cite{Li2019dataset}          & 2019 & deepfake                    &        -- / 5,639    & various                           & MPEG4           \\ \hline
		\ru DFDC-preview     & \cite{Deepfake_FB}            & 2019 & deepfake                &      1,131 / 4,113    & 180p $-$ 2160p                 & H.264            \\ \hline
		\ru DFDC     & \cite{Deepfake_Kaggle}            & 2019 & deepfake                &      19,154 / 100,000    & 240p $-$ 2160p                 & H.264            \\ \hline
		\ru DeeperForensics-1.0     & \cite{Jiang2020}   & 2020 & deepfake                &      50,000 / 10,000    & 1080p                 & --           \\ \hline
	\end{tabular}
	\label{tab:datasets_videos}
\end{table*}

In \cite{Korshunov2018deepfakes} a face-swapping video dataset, DF-TIMIT, has been built, with 620 deepfake videos obtained with a GAN-based approach.
The original data come from a database which contains 10 videos for each of 43 subjects.
16 couples of subjects were manually chosen from the database in order to generate videos with swapped faces from subject one to subject two and viceversa,
producing both a low quality and a high quality video.
In \cite{Khodabakhsh2018}, instead, proposes the Fake Face in the Wild Dataset, FFW,
comprising only 150 manipulated videos which, however, show a large variety of approaches, including splicing and CG faces, using both manual effort and completely automatic procedures.
Finally, in \cite{Papadopoulou2018}, manipulated videos retrieved from the web have been collected in a dataset that includes 200 fake videos and 180 real videos.
An extended version of this dataset also presents near-duplicates found on the web.

\begin{figure}[t!]
	\centering
	\begin{tabular}{cc}
		\includegraphics[width=1.75\linewidth, trim=10 230 0 0, clip]{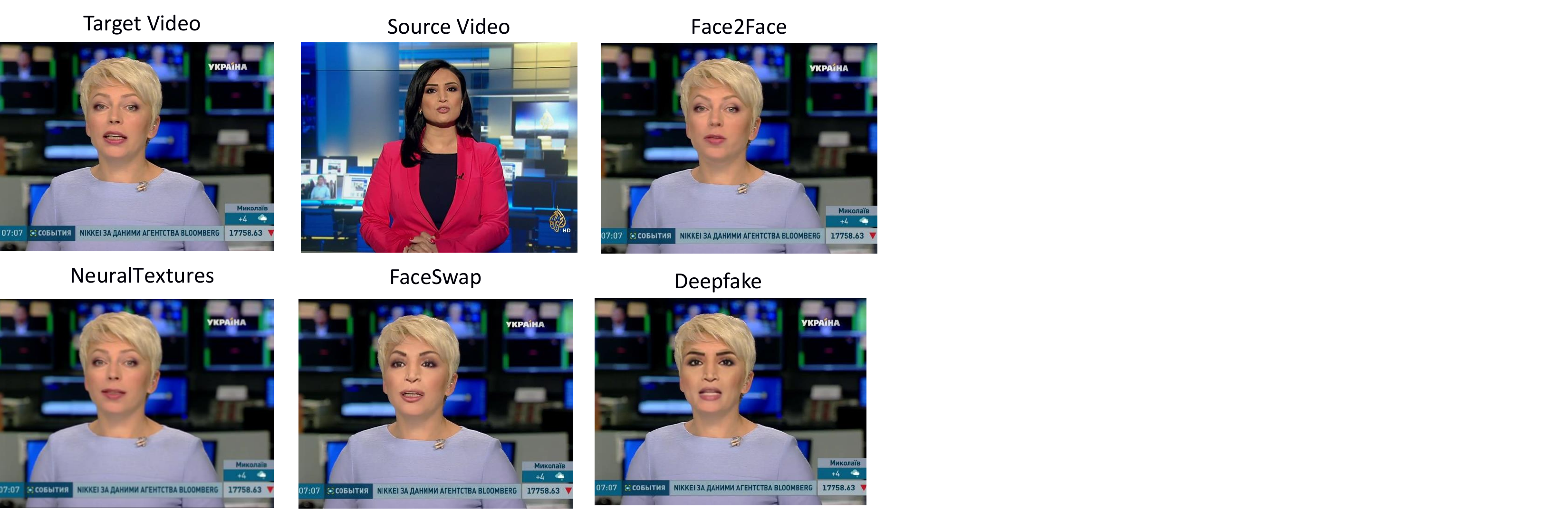}
	\end{tabular}
	\caption{Example manipulated videos from FaceForensics++.
		A single original video (top-left) is manipulated by four different tools (Face2Face, NeuralTextures, FaceSwap, DeepFake)
		using information drawn from a different source video.}
	\label{fig:FF}
\end{figure}

The first large dataset with automatically manipulated faces, FaceForensics++, has been proposed in \cite{Roessler2019}.
It contains 1,000 original videos downloaded from the YouTube-8M dataset \cite{YouTube8M} and 4,000 manipulated videos obtained from them by using four different manipulation tools.
Two of them are based on computer-graphics and two on deep learning, two perform changes of expression and two face swapping,
Figure \ref{fig:FF} shows a few examples.
The dataset is available in uncompressed and H264 compressed format, with two quality levels, in order to stimulate developing methods robust to compression.
Recently, Google and Jigsaw contributed the dataset with 3,000 more manipulated videos, created {\it ad hoc} using 28 actors \cite{Dufour2019}.
Also in \cite{Li2019dataset} a new deepfake video dataset has been introduced, called Celeb-DF.
It comprises 5,639 manipulated videos,
the real videos are based on publicly available YouTube video clips of 59 celebrities of diverse genders, ages, and ethnic groups.
Forged videos are created by swapping faces for each pair of the 59 subjects using an improved deepfake synthesis method.
Instead in \cite{Deepfake_FB} the first release of the dataset used for the Facebook DeepFake Detection Challenge (DFDC) is described.
It is composed by 4,113 deepfake videos created using two different synthesis algorithms on the basis of 1,131 original videos featuring 66 enrolled actors.
The final dataset made available for the Kaggle competition \cite{Deepfake_Kaggle} (started on December 2019)
is instead much larger. It comprises 100,000 manipulated videos and around 19,000 pristine ones. 
A very recent dataset has been built in \cite{Jiang2020}, comprising 10,000 fake videos built using 100 actors and
applying 7 perturbations, like color saturation, blurring and compression,  
with different parameters for a total of 35 possible post-processing so as to better represent a real scenario.

\subsection{Original media}
	
Since some methods work on anomaly detection and are trained only on pristine data,
it makes sense to describe also a list of useful publicly available resources based on datasets of authentic images and videos.
Of course, they can also be used to simulate different types of manipulation and carry out initial experiments on a synthetic dataset.
They are commonly used for source/camera identification, but this task is strictly related to media forgery detection.
For example, in case of a composition,
by identifying the origin of image patches, one can detect the presence of multiple sources.

The Dresden image database \cite{Gloe2010dresden} is the most popular one.
It contains over 14,000 JPEG images captured by 73 digital cameras of 25 different models.
Raw images, instead, can be found in the RAISE dataset \cite{Nguyen2015raise}, composed by 8,156 images taken by 3 different cameras.
A further dataset was released for the 2018 Kaggle competition \cite{Kaggle} on camera model identification and is also available on-line.
It is composed by 2,750 images from 10 different camera models.
A dataset that contains SDR (Standard Dynamic Range) and 
HDR (High Dynamic Range) images has been presented in \cite{Shaya2018}.
A total of 5,415 images were captured by 23 mobile devices
in different scenarios under controlled acquisition conditions.
The VISION dataset instead \cite{Shullani2017} includes 34,427 images and 1,914 videos from 35 devices of 11 major brands.
Media assets are both in their original format and as they appear after uploading/downloding on various platforms (Facebook, YouTube, WhatsApp)
so as to allow studies on data downloaded from social networks.
Another mixed dataset, SOCRATES, is proposed in \cite{Galdi2017}.
It contains 6,200 images and 680 videos captured using 67 smartphones of 14 brands and 42 models.
Finally, this year the video-ACID database has been published \cite{Hosler2019}, with over 12,000 videos from 46 physical devices of 36 different models.

%% file: counterforensics.tex
\section{Counterforensics}

In multimedia forensics, like in other security-related fields,
one should always account for the presence of an adversary which actively tries to mislead the analyses.
In fact, a skilled attacker, aware of the principles on which forensic tools rely,
may enact a number of counter-forensic measures on purpose to evade detectors \cite{Gloe2007}.
Forensic tools should prove robust to such attacks,
as well as to all real-world conditions that tend to impair the performance observed in laboratory.
Therefore, the many counterforensics methods designed to fool current detectors represent a precious help towards the development of multimedia forensics,
since they highlight the weaknesses of current solutions and stimulate research for more robust ones \cite{Bohme2012}.

A large body of literature concerns attacks targeted to specific forensic methods, which try to exploit their weaknesses.
For example, some methods try to hide traces of resampling that manifest as strong peaks in the Fourier domain \cite{Kirchner2008hiding}.
Also sensor traces, and especially PRNU fingerprints, are popular targets because of their importance in forensics.
So, methods have been proposed to remove the true device fingerprint from an image,
and also to inject the fingerprint of a different device in it \cite{Gloe2007}.
Besides hindering source identification, these attacks can reduce the ability to discriminate manipulated from pristine data.
As said before, however, they stimulated the design of more robust detectors \cite{Goljan2011},
and then more powerful attacks \cite{Marra2014}, in an arms race typical of such two-player games.
Attacks to traditional methods will not be explored, referring the reader to a very recent review by Barni {\it et al.} \cite{Barni2018},
and focus instead on counterforensics for deep learning-based methods.

\begin{figure}[t!]
	\centering
	\begin{tabular}{cc}
		\includegraphics[width=1.1 \linewidth, trim=50 180 0 100, clip]{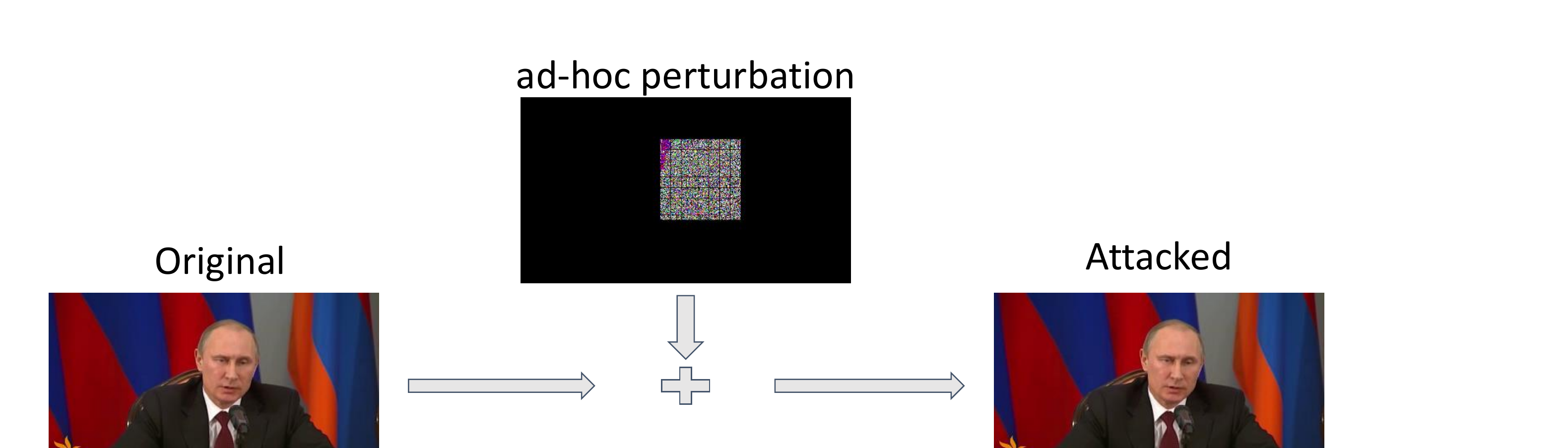}
	\end{tabular}
	\caption{Fooling deepfake detectors. A suitable inconspicuous adversarial noise pattern can be added to a deepfake video 
		to mislead the CNN detector into classifying it as pristine.}
	\label{fig:attack}
\end{figure}

Back in 2014, in the context of image recognition, it has been shown \cite{Szegedy2014} that
convolutional neural networks are extremely vulnerable to adversarial attacks.
By injecting a suitable inconspicuous noise pattern in the target image, 
the attacker can induce the network to change classification in any desired way (see Figure \ref{fig:attack}).
Actually, this is not a new problem, the vulnerability of machine learning had been known for several years, 
especially in applications like spam and malware detection,
and has been thoroughly studied \cite{Biggio2018} in the adversarial machine learning field.
In any case, these alarming findings have spawned intense research on this topic in computer vision.

In multimedia forensics,
some early papers \cite{Marra2015, Chen2017} focused on attacking machine learning detectors based on the rich model handcrafted features, 
obtaining mixed results with high complexity.
Indeed, attacking CNNs seems to be easier and more effective \cite{Guera2017}.
Backpropagation provides a strong help to design gradient-based adversarial perturbations.
However, differently from what happens in computer vision \cite{Goodfellow2015},
it appears that adversarial noise designed to fool a specific CNN architecture does not 
transfer to different architectures trained for the same task \cite{Gragnaniello2018, Barni2019}.
This is probably because the adversarial noise lives in the same space (high frequencies) where major forensic clues live.
Lossy compression is a further intrinsic defence against adversarial attacks \cite{Das2018}.
Indeed, strong lossy compression, ubiquitous in real-world scenarios,
removes not only useful forensic traces but also adversarial noise, reducing the effectiveness of such attacks \cite{Marra2018vulnerability}.
Table \ref{fig:attacks} shows some experimental results on Face2Face manipulated videos \cite{Roessler2019}.
Attacks crafted with FSGM \cite{Goodfellow2015} with various strengths ($\epsilon=1,2,3$)
have been applied only on faces  (so as to not distort the visual quality, Figure \ref{fig:attack})
and the performance of a CNN-based detector are evaluated.
With $\epsilon=3$ the detector accuracy becomes close to 50\% (random choice).
However, if videos are compressed, a large part of the adversarial noise is removed, and the detector performance improves again.
 
Recently, some methods have been specifically developed to attack CNN detectors for multimedia forensics applications.
In \cite{Kim2018sps} a GAN-based architecture is proposed to hide the traces of median filtering in order to fool state-of-the-art CNN-based detectors.
A different perspective is followed in \cite{Chen2019},
where a generative method has been proposed to falsify the camera model traces.
The attack does not only fool camera model classifiers, but reduces also the power of forensic analyses based on traces of in-camera processing.
Along this same direction,
in \cite{Neves2019} an autoencoder-based method is proposed to remove GAN fingerprints and impair the performance of systems designed to detect GAN-generated images.
Instead \cite{Cozzolino2019spoc} proposes a GAN-based architecture with a twofold goal,
to inject traces of a real camera in synthetic images and, at the same time, reduce peculiar traces of GAN generation.
Therefore, attacked images cannot be recognized anymore as computer-generated and are instead recognized as real images of the target camera.
The attack takes place in a black-box scenario, with no information on the attacked detectors.

Notably, all these approaches preserve a very good image quality, with no perceivable visual artifacts,
demonstrating the urgent need of stronger and more robust detection methods. 

%% file: fusion.tex
\begin{table}[t!]
	{\footnotesize
		\caption{Results on Face2Face manipulations in the presence of adversarial noise and compression}
		\label{fig:attacks}
		\centering
		\begin{tabular}{l||c|c|c|} \cline{2-4}
			\ru                 & \multicolumn{3}{c|}{Accuracy} \\ \cline{2-4}
			\ru                 & Uncompressed & High Quality & Low Quality \\ \hline\hline
			\ru  no attack      & 99.93\% & 98.13\%  & 87.81\% \\ \hline
			\ru  low attack     & 80.43\% &  94.83\% & 85.83\% \\ \hline
			\ru  medium attack  & 56.37\% &  89.93\% & 83.30\% \\ \hline
			\ru  strong attack  & 52.23\% &  82.00\% & 80.30\% \\ \hline
		\end{tabular}
	}
\end{table}

\section{Fusion}

A system with the ambition to provide reliable decisions about the integrity of images and videos
must necessarily integrate multiple tools, to cover most, if not all, the operating conditions of interest.
In fact, each individual method works under suitable hypotheses, and can become completely useless when these do not hold.
For example, a tool for copy-move detection will not help in case of a splicing.
Fusing multiple tools, however, is not only important to widen the spectrum of detectable forgeries,
but also to improve the detection capability with respect to each single one.
In fact, the traces to be detected are usually extremely weak,
and can be easily masked by intentional attacks, as described in the previous Section, and even by standard processing steps.
Therefore, the integration of multiple tools designed to detect similar attacks with different approaches,
may be expected to improve performance, and especially robustness, to both innocent and malicious disturbances.
On the other hand, maximizing the number of clues is standard practice in investigative procedures.

\begin{figure}[t!]
	\centering
	\begin{tabular}{cc}
		\includegraphics[width=1.35\linewidth, trim=30 180 0 100, clip]{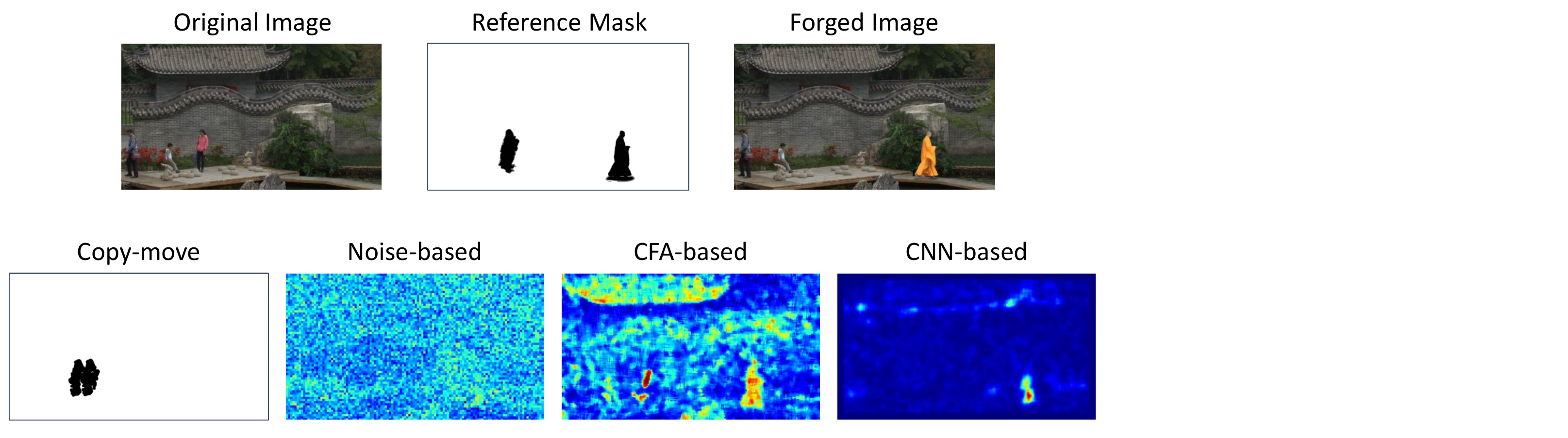}
	\end{tabular}
	\caption{Using information fusion to detect multiple manipulations.
    Different tools provide different pieces of information, which may locally agree, disagree, or being complementary.
    A human expert can likely make sense of all such clues, but transferring this expertise to a computer may be difficult.}
	\label{fig:fusion}
\end{figure}

\begin{figure*}[t!]
	\centering
	\begin{tabular}{cc}
		\includegraphics[width=1.20\linewidth, trim=100 320 0 40, clip]{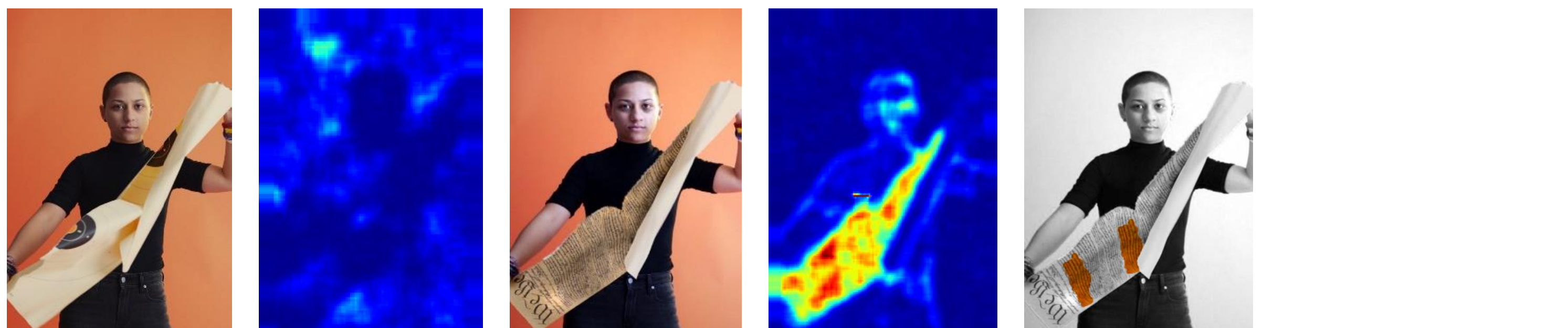}
	\end{tabular}
	\caption{From left to right: the pristine Emma Gonzalez image, the heatmap obtained using a CNN-based localization method, the manipulated image, and the corresponding heatmap obtained with the same CNN-based detector. The last image shows the output of a copy-move detector operating on the manipulated version. Therefore, multiple clues
	were found: splicing and cloning. Also in this case deciding the right fusion strategy is not easy.}
	\label{fig:Gonzalez}
\end{figure*}

The typical workflow of a forensic tool is to extract suitable low-level features from the original data,
process them to obtain a scalar score or a probability vector,
and eventually process the latter (score thresholding, max probability choice) to make the final decision.
Fusion may take place at all three levels, called feature, measurement, and abstract level, with pros and cons.
As observed in \cite{Fontani2013},
working at the feature level presents serious drawbacks when a large number of tools are involved,
especially for the number of features to deal with, and the problem of creating representative datasets for all possible situations of interest.
On the other hand, with abstract-level fusion,
precious information may have been already discarded, reducing the ability to exploit cross-tool dependencies.
Working at the measurement level may be a reasonable compromise in between these two extremes.
In any case,
while fusion is certainly a major asset to improve performance,
it is not obvious how to combine wildly different pieces of information in a sensible way, as shown by the example of Figure \ref{fig:fusion}.
Here, an expert can easily understand that manipulations occurred, and where, but transferring such a skill to a machine is not trivial.
Another example is shown in Figure \ref{fig:Gonzalez},
where the two versions of Emma Gonzalez are analyzed using a 
CNN-based localization approach. For the first image the heatmap shows that there is no manipulation,
while for the second image the forged area is highlighted.
What is interesting is that a further manipulation was carried out by the malicious user,
who also replicated some text inside the poster, which is detected by a copy-move based detector.
Indeed, to create a realistic forgeries multiple modifications were needed and hence multiple traces were left.
How to fuse these results is clearly not trivial.

Despite the intense research activity in multimedia forensics,
there has been limited attention on the fusion of diverse and complementary tools.
One of the first contributes is the study of Bayram {\it et al.} \cite{Bayram2006} back in 2006.
Their results clearly show that detectors based on fusion dominate all individual base detectors in terms of accuracy,
and the gain is more significant when operating at the abstract level.
In \cite{Fontani2013, Ferrara2015} fusion is addressed in a more systematic way,
relying on the Dempster-Shafer theory (DST) of evidence \cite{Dempster1967} to meaningfully combine multiple tools at the measurement level.
The DST provides a methodology to include concepts like uncertainty, reliability, and compatibility in the decision process.
In fact, to attribute the correct importance to a detector score,
one should take into account its overall reliability, 
the level of confidence associated with the specific decision and its compatibility with other detectors. 
Experiments in \cite{Fontani2013, Ferrara2015} show that the DST approach, with reasonable fusion rules,
outperforms consistently all individual tools and also abstract-level fusion, thanks to the use of richer information.
It even outperforms machine-learning fusion in the absence of a training set well-aligned with the test set,
a recurrent situation in multimedia forensics.
In \cite{Cozzolino2013} several forensic tools, based on complementary hypotheses, are fused following various strategies,
with results confirming that measurement-level fusion is more effective than abstract-level fusion.
In \cite{Korus2016a}, instead, contextual information is taken into account by means of prior Markovian models,
while in \cite{Korus2016} multi-scale fusion is investigated to improve the localization accuracy of tools operating in sliding-window modality.

It is worth mentioning that fusion is a common characteristic of all the winning approaches in several forensic challenges,
like in the IEEE Forensics challenge \cite{Cozzolino2014a, Cozzolino2014b},
the fraud detection contest on receipts \cite{Artaud2018}
or the Camera Model Identification Challenge organized for the 2018 IEEE Signal Processing Cup and hosted on Kaggle \cite{Kaggle}.
More in general, in the data science community,
it is standard practice to combine the probability vectors output by multiple deep networks, even by simple averaging.
Empirically, this fusion improves robustness when moving from training to test data,
more and more with the increasing number of networks, in line with the ``wisdom of the crowd'' principle.

%% file: future_work.tex
\section{Future work}

As clearly shown by this survey, in the last fifteen years there has been intense research on multimedia forensics, and great progresses.
Nonetheless, many issues remain unsolved,
new challenges appear by the day and, eventually, most of the road seems to be still ahead of us.
This is not so surprising, though.
For sure, the advent of deep learning has given extraordinary impulse to both media manipulation methods and forensic tools, opening new research areas.
A more fundamental reason, however, is the two-player nature of this research field.
The presence of skilled attackers is a guarantee that
no tool will protect us forever, and new solutions will be always necessary to cope with unforeseen menaces.
With this premise, it is important trying to identify the most promising areas for future research.

A first topic is fusion.
As manipulations get smarter and smarter, individual tools will become ever less effective against a wide variety of attacks.
Therefore, multiple detection tools, multiple networks, multiple approaches must be put to work together,
and how to best combine all available pieces of information should be the object of a more sustained research.
Besides multi-tool fusion, also multi-asset analysis should be pursued.
More and more, the individual media assets should be analyzed together with all correlated evidence.
A picture or a video used to spread a fake news should not be studied in isolation but
together with the accompanying text \cite{Dhruv2019}, audio \cite{Korshunov2019},
and all available contextual information \cite{Boididou2017}.
Also, the approach can be changed based on the availability of additional information, {\it e.g.}, metadata or near-identical versions of the image/video under test.
Eventually, a whole array of semantic-level analyses should be pursued,
as envisioned by the recent initiative launched by DARPA on Semantics Forensics.

Focusing more specifically on deep learning-based tools,
the main technical issue is probably the (in)ability of deep networks to adapt to situations not seen in the training phase.
This issue emerges in several circumstances.
First of all, media assets undergo a number of innocent processing steps, like compression, resizing, rotation, re-capturing, etc.,
that modify significantly their high-order statistics, so precious for forgery detection.
Besides innocent transformations, malicious ones should be also considered, designed specifically to disguise forensic clues.
It is unlikely that all combinations of such transformations can be represented in a training set.
Therefore, higher robustness should be pursued by other means.
Also, to deal with the rapid advances in manipulation technology, deep networks should be able to adapt readily to new manipulations,
without a full re-training, which may be simply impossible for lack of training data or entail a catastrophic forgetting phenomena.

Another hot issue for deep learning-based methods is interpretability.
The black-box nature of deep learning makes it difficult to understand why a certain decision is made.
A deep network may correctly classify a cat's picture as ``cat'', but we do not know exactly which specific features motivated this decision.
Of course, this is a serious issue for some forensic applications.
For example, a judge would hardly base decisions only on statistical bases.
More in general, being able to track the reasoning of a deep network would allow to improve its design and training phase,
and provide higher robustness with respect to malicious attacks.

Lastly,
we underline a resurgent of interest on active authentication methods \cite{Singh2019, Zheng2020}.
In past decades, a large body of research was produced on digital watermarking \cite{Cox2008}.
There is now much interest in blockchain technology \cite{Hasan2019}, in criptography \cite{Boneh2019},
and even new active methods have been proposed to ensure the integrity of digital media \cite{Korus2019}
or to protect individuals from becoming the victims of AI attacks \cite{Li2019hiding}.
As we said before, despite its long history, multimedia forensics appears to be still in full development,
with high demands from industry and society and many answers yet to be given.